%% file: neurips_data_2023.tex
\documentclass{article}

\usepackage[final]{neurips_data_2023}
\usepackage{graphicx}
\usepackage{subcaption}

\usepackage[utf8]{inputenc} %
\usepackage[T1]{fontenc}    %
\usepackage{hyperref}       %
\usepackage{url}            %
\usepackage{booktabs}       %
\usepackage{amsfonts}       %
\usepackage{nicefrac}       %
\usepackage{microtype}      %
\usepackage{xcolor}         %

\usepackage{booktabs}
\usepackage{multirow}
\usepackage{cleveref}
\usepackage{enumitem}
\usepackage{listings}
\usepackage{float}

\setitemize{noitemsep,topsep=0pt,parsep=0pt,partopsep=0pt,leftmargin=1.5em}

\newcommand{\showcomments}{yes}

\newcommand\todo[1]{\ifthenelse{\equal{\showcomments}{yes}}{{\color{red} TODO: #1}}{\ignorespaces}}
\newcommand\joey[1]{\ifthenelse{\equal{\showcomments}{yes}}{{\color{blue} (Joey: #1)}}{\ignorespaces}}
\newcommand\lianmin[1]{\ifthenelse{\equal{\showcomments}{yes}}{{\color{blue} Lianmin: #1}}{\ignorespaces}}
\newcommand\ion[1]{\ifthenelse{\equal{\showcomments}{yes}}{{\color{blue} Ion: #1}}{\ignorespaces}}
\newcommand\hao[1]{\ifthenelse{\equal{\showcomments}{yes}}{{\color{blue} Hao: #1}}{\ignorespaces}}
\newcommand\dacheng[1]{\ifthenelse{\equal{\showcomments}{yes}}{{\color{blue} Dacheng: #1}}{\ignorespaces}}
\newcommand\weilin[1]{\ifthenelse{\equal{\showcomments}{yes}}{{\color{cyan} Wei-Lin: #1}}{\ignorespaces}}
\newcommand\siyuan[1]{\ifthenelse{\equal{\showcomments}{yes}}{{\color{purple} Siyuan: #1}}{\ignorespaces}}

\title{Judging LLM-as-a-Judge \\ with MT-Bench and Chatbot Arena}

\author{
Lianmin Zheng$^1$\thanks{Joint first authors. This paper is an extended version of our earlier blog post~\cite{vicuna2023}.} $\quad$ Wei-Lin Chiang$^{1*}$ $\quad$ Ying Sheng$^{4*}$ $\quad$ Siyuan Zhuang$^{1}$
\And
Zhanghao Wu$^1$ $\quad$ Yonghao Zhuang$^3$ $\quad$ Zi Lin$^2$ $\quad$ Zhuohan Li$^1$ $\quad$ Dacheng Li$^{13}$
\And Eric P. Xing$^{35}$ $\quad$ Hao Zhang$^{12}$ \quad  Joseph E. Gonzalez$^1$ $\quad$ Ion Stoica$^1$ \\\\
$^1$ UC Berkeley \quad $^2$ UC San Diego \quad $^3$ Carnegie Mellon University \quad $^4$ Stanford \quad $^5$ MBZUAI
}
\begin{document}

\maketitle
    
\input{sec-intro}

\input{sec-mt-bench}

\input{sec-llm-judge}

\input{sec-exp}
\input{sec-train-data}

\input{sec-discussion}

\bibliographystyle{plain}
\bibliography{reference}

\newpage
\appendix

\input{appendix}

\end{document}

%% file: sec-intro.tex
\begin{abstract}
Evaluating large language model (LLM) based chat assistants is challenging due to their broad capabilities and the inadequacy of existing benchmarks in measuring human preferences.
To address this, we explore using strong LLMs as judges to evaluate these models on more open-ended questions.
We examine the usage and limitations of LLM-as-a-judge, including position, verbosity, and self-enhancement biases, as well as limited reasoning ability, and propose solutions to mitigate some of them.
We then verify the agreement between LLM judges and human preferences by introducing two benchmarks: MT-bench, a multi-turn question set; and Chatbot Arena, a crowdsourced battle platform.
Our results reveal that strong LLM judges like GPT-4 can match both controlled and crowdsourced human preferences well, achieving over 80\% agreement, the same level of agreement between humans.
Hence, LLM-as-a-judge is a scalable and explainable way to approximate human preferences, which are otherwise very expensive to obtain.
Additionally, we show our benchmark and traditional benchmarks complement each other by evaluating several variants of LLaMA and Vicuna.
The MT-bench questions, 3K expert votes, and 30K conversations with human preferences are publicly available at \url{https://github.com/lm-sys/FastChat/tree/main/fastchat/llm\_judge}.
\end{abstract}

\section{Introduction}
\label{sec:introduction}
There has been a proliferation of LLM-based chat assistants (chatbots) that leverage supervised instruction fine-tuning and reinforcement learning with human feedback (RLHF) to unlock new instruction following and conversational abilities~\cite{ouyang2022training,bai2022training,openai2023gpt4,vicuna2023,zhou2023lima,xu2023wizardlm,dubois2023alpacafarm}.
Once aligned with humans, these chat models are strongly preferred by human users over the original, unaligned models on which they are built.
However, the heightened user preference does not always correspond to improved scores on traditional LLM benchmarks -- benchmarks like MMLU~\cite{hendrycks2020measuring} and HELM~\cite{liang2022holistic} cannot effectively tell the difference between these aligned models and the base models.
This phenomenon suggests that there is a fundamental discrepancy between user perceptions of the usefulness of chatbots and the criteria adopted by conventional benchmarks.

\begin{figure}[t]
    \centering
    \vspace{-1em}
    \includegraphics[width=0.95\linewidth]{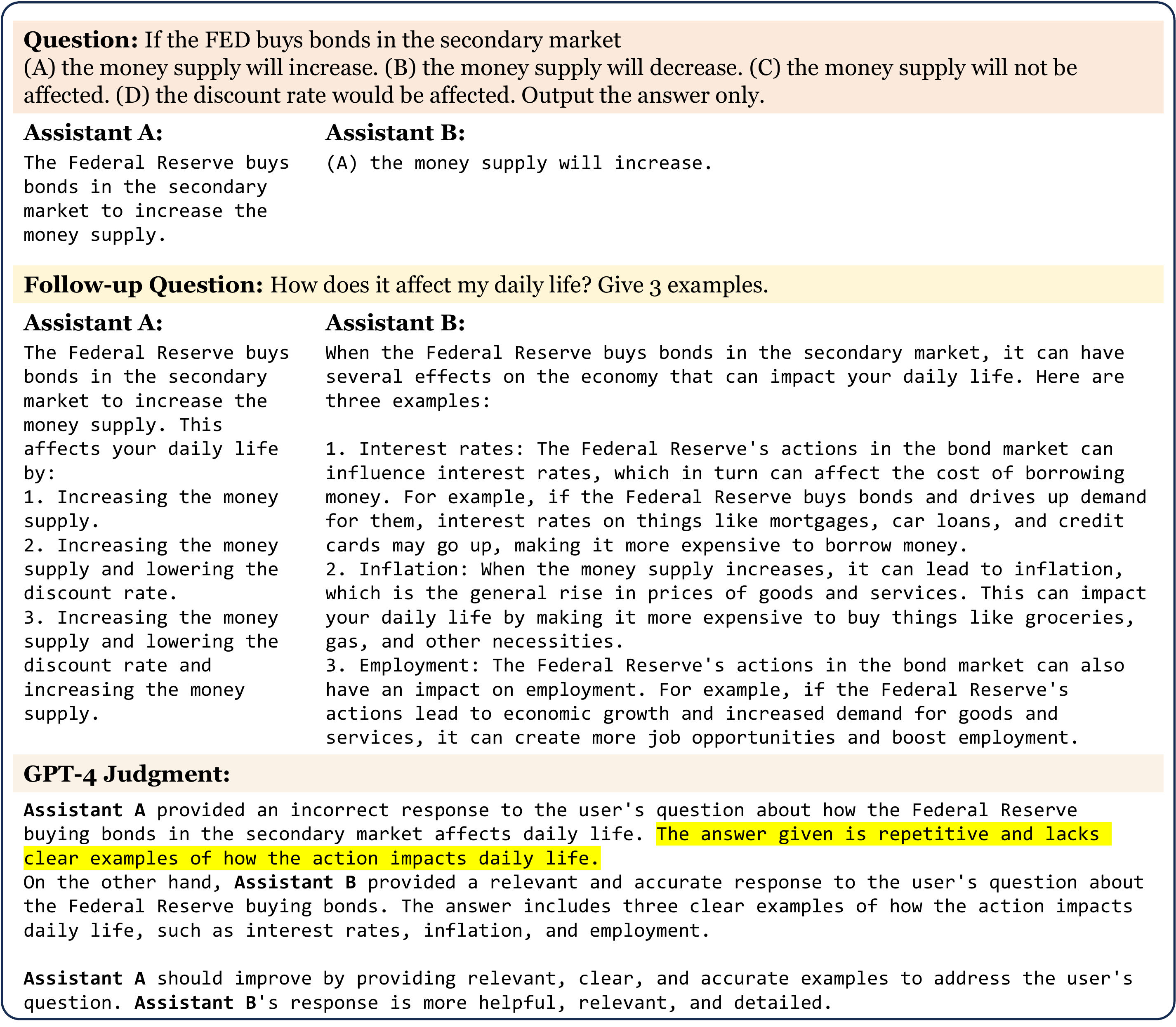}
    \vspace{-0.5em}
    \caption{Multi-turn dialogues between a user and two AI assistants—LLaMA-13B (Assistant A) and Vicuna-13B (Assistant B)—initiated by a question from the MMLU benchmark and a follow-up instruction. GPT-4 is then presented with the context to determine which assistant answers better.}
    \label{fig:intro-figure}
    \vspace{-1.5em}
\end{figure}

We argue that this discrepancy primarily arises due to  
existing evaluation that only measures LLMs' core capability on a confined set of tasks (e.g., multi-choice knowledge or retrieval questions), without adequately assessing its alignment with human preference in open-ended tasks, such as the ability to accurately adhere to instructions in multi-turn dialogues.
As a demonstration, we show conversation histories with two models on an MMLU question in \Cref{fig:intro-figure}.
The two models are LLaMA-13B~\cite{touvron2023llama}, a pre-trained base model without fine-tuning, and Vicuna-13B, our fine-tuned model from LLaMA-13B on high-quality conversations (the training details are in \Cref{sec:training-details}).
Despite the base LLaMA models showing competitive performance on conventional benchmarks (\Cref{table:dataset_composition}), its answers to open-ended questions are often not preferred by humans.
This misalignment of conventional benchmarks underscores the core problem driving this paper:
\emph{the need for a robust and scalable automated method to evaluate LLM alignment with human preferences.}

To study this, we introduce two benchmarks with human ratings as the primary evaluation metric: MT-bench and Chatbot Arena.
MT-bench is a series of open-ended questions that evaluate a chatbot's multi-turn conversational and instruction-following ability -- two critical elements for human preference. MT-bench is also carefully constructed to differentiate chatbots based on their core capabilities, such as reasoning and math.
In addition, we develop Chatbot Arena, a crowdsourced platform featuring anonymous battles between chatbots in real-world scenarios -- Users engage in conversations with two chatbots at the same time and rate their responses based on personal preferences.

While human evaluation is the gold standard for assessing human preferences, it is exceptionally slow and costly.
To automate the evaluation, we explore the use of state-of-the-art LLMs, such as GPT-4, as a surrogate for humans. 
Because these models are often trained with RLHF, they already exhibit strong human alignment.
We call this approach \emph{``LLM-as-a-judge''}. 
This approach has been tried in our earlier blog post~\cite{vicuna2023} and other concurrent or follow-up work~\cite{bubeck2023sparks, openaievals, dubois2023alpacafarm, dettmers2023qlora, zhou2023lima,gudibande2023false,peng2023instruction,wang2023large,chiang2023can,wang2023far}.
However, there has not been a systematic study of this approach. %

In this paper, we study the LLM-as-a-judge approach by comparing it to the gold standard of human evaluation. 
We examine several potential limitations of the LLM-as-a-judge approach including position bias, verbosity bias, self-enhancement bias, and limited reasoning ability.
We show that some of the biases are minor or can be mitigated. 
Once addressed, our results from 3K controlled expert votes and 3K crowdsourced human votes in the wild verify that GPT-4 judge match human evaluations at an agreement rate exceeding 80\%, achieving the same level of human-human agreement (\S\ref{sec:agreement}, Table~\ref{tab:failure_rate_math}).
Consequently, this suggests LLM-as-a-judge is a scalable method to swiftly evaluate human preference, serving as a promising alternative to traditional human evaluations.  \\

This paper makes two contributions: (1) a systematic study of LLM-as-a-judge; and (2) human preference datasets with high-quality questions and diverse user interactions from MT-bench and Chatbot Arena.
In addition, we argue for the adoption of a hybrid evaluation framework for future LLM benchmarks: by combining the existing capability-based benchmarks and the new preference-based benchmarks with LLM-as-a-judge, one can swiftly and automatically evaluate both the core capabilities and human alignment of models.
We publicly release 80 MT-bench questions, 3K expert votes, and 30K conversations with human preferences for future study.

%% file: sec-mt-bench.tex
\begin{table}[H]
\footnotesize
\vspace{-1em}
\caption{Sample multi-turn questions in MT-bench. }
\label{tab:mt_bench_sample}
\centering
\resizebox{0.9\columnwidth}{!}{
\begin{tabular}{@{}p{0.10\linewidth}|p{0.08\linewidth}p{0.75\linewidth}@{}}
\toprule
Category                  & \multicolumn{2}{c}{Sample Questions}                                                                                                                                                                                                                                                                           \\ \midrule
\multirow{2}{*}{Writing}  & 1st Turn & Compose an engaging travel blog post about a recent trip to Hawaii, highlighting cultural experiences and must-see attractions.                                                                                                                                                                        \\ \cmidrule(l){2-3} 
                          & 2nd Turn & Rewrite your previous response. Start every sentence with the letter A.                                                                                                                                                                                                                                \\ \midrule
\multirow{2}{*}{Math}     & 1st Turn & Given that $f(x) = 4x^3 - 9x - 14$, find the value of $f(2)$. \\ \cmidrule(l){2-3} 
                          & 2nd Turn & Find $x$ such that $f(x) = 0$. \\ \midrule
\multirow{2}{*}{Knowledge} & 1st Turn & Provide insights into the correlation between economic indicators such as GDP, inflation, and unemployment rates. Explain how fiscal and monetary policies ...                                                                                                                   \\ \cmidrule(l){2-3} 
                          & 2nd Turn & Now, explain them again like I’m five.                                                                                                                                                                                                                                                                 \\ \bottomrule
\end{tabular}
}
\vspace{-0.5em}
\end{table}

\section{MT-Bench and Chatbot Arena}
\label{sec:mt-bench}

\subsection{Motivation}
With the recent advances of LLMs, LLM-based assistants start to exhibit artificial general intelligence across diverse tasks, from writing and chatting to coding~\cite{bubeck2023sparks,openai2023gpt4,anil2023palm,srivastava2022beyond}. However, evaluating their broad capabilities also becomes more challenging. 
Despite the availability of numerous benchmarks for language models, they primarily focus on evaluating models on closed-ended questions with short responses. 
Given that these chat assistants can now precisely follow user instructions in multi-turn dialogues and answer open-ended questions in a zero-shot manner, current benchmarks are inadequate for assessing such capabilities. Existing benchmarks mostly fall into the following three categories.
\begin{itemize}
\item \textbf{Core-knowledge benchmarks}, including MMLU~\cite{hendrycks2020measuring}, HellaSwag~\cite{zellers2019hellaswag}, ARC~\cite{clark2018think}, WinoGrande~\cite{sakaguchi2021winogrande}, HumanEval~\cite{chen2021evaluating}, GSM-8K~\cite{cobbe2021training}, and AGIEval~\cite{zhong2023agieval}, evaluate the core capabilities of pre-trained LLMs using zero-shot and few-shot benchmark sets. They typically require LLMs to generate a short, specific answer to benchmark questions that can be automatically validated. %
\item \textbf{Instruction-following benchmarks}, such as  Flan~\cite{longpre2023flan,wei2021finetuned}, Self-instruct~\cite{selfinstruct}, NaturalInstructions~\cite{naturalinstructions}, Super-NaturalInstructions~\cite{supernaturalinstructions}, expand to slightly more open-ended questions and more diverse tasks and are used to evaluate LLMs after instruction fine-tuning.
\item \textbf{Conversational benchmarks}, like CoQA~\cite{reddy2019coqa}, MMDialog~\cite{feng2022mmdialog} and OpenAssistant~\cite{kopf2023openassistant}, are closest to our intended use cases. However, the diversity and complexity of their questions often fall short in challenging the capabilities of the latest chatbots.
\end{itemize}

While largely overlooked by existing LLM benchmarks, human preferences serve as a direct measure of a chatbot's utility in open-ended, multi-turn human-AI interactions. To bridge this gap, we introduce two novel benchmarks expressly tailored to assess human preferences. Simultaneously, these benchmarks are designed to distinguish the core capabilities of state-of-the-art models.

\subsection{MT-Bench}
We create MT-bench, a benchmark consisting of 80 high-quality multi-turn questions.
MT-bench is designed to test multi-turn conversation and instruction-following ability, covering common use cases and focusing on challenging questions to differentiate models.
We identify 8 common categories of user prompts to guide its construction: writing, roleplay, extraction, reasoning, math, coding, knowledge I (STEM), and knowledge II (humanities/social science).
For each category, we then manually designed 10 multi-turn questions.
\Cref{tab:mt_bench_sample} lists several sample questions.

\subsection{Chatbot Arena}
Our second approach is Chatbot Arena, a crowdsourcing benchmark platform featuring anonymous battles.
On this platform, users can interact with two anonymous models simultaneously, posing the same question to both. They vote for which model provides the preferred response, with the identities of the models disclosed post-voting. After running Chatbot Arena for one month, we have collected around 30K votes. Since the platform does not use pre-defined questions, it allows gathering a wide range of unrestricted use cases and votes in the wild, based on the diverse interests of users. A screenshot of the platform can be found at \Cref{subsec:chatbot-arena}.

%% file: sec-llm-judge.tex
\section{LLM as a Judge}
\label{sec:llm-as-a-judge}
While our initial evaluations using MT-bench and Chatbot Arena rely on human ratings, collecting human preferences can be costly and laborious~\cite{selfinstruct,alpaca,ouyang2022training,bai2022training,diao2023lmflow}. To overcome this, we aim to develop a more scalable and automated approach. Given that most questions in MT-bench and Chatbot Arena are open-ended without reference answers, devising a rule-based program to assess the outputs is extremely challenging.
Traditional evaluation metrics based on the similarity between outputs and reference answers (e.g., ROUGE~\cite{lin2004rouge}, BLEU~\cite{papineni2002bleu}) are also ineffective for these questions.

As LLMs continue to improve, they show potential in replacing human annotators in many tasks~\cite{gilardi2023chatgpt,huang2023chatgpt}. Specifically, we are interested in whether LLMs can effectively evaluate the responses of chat assistants and match human preferences. Next, we discuss the use and limitations of LLM-as-a-judge.

\subsection{Types of LLM-as-a-Judge}
We propose 3 LLM-as-a-judge variations. They can be implemented independently or in combination:
\begin{itemize}
\item \textbf{Pairwise comparison}. An LLM judge is presented with a question and two answers, and tasked to determine which one is better or declare a tie. The prompt used is given in \Cref{fig:default-prompt-pair} (Appendix).
\item \textbf{Single answer grading}.
Alternatively, an LLM judge is asked to directly assign a score to a single answer. The prompt used for this scenario is in \Cref{fig:default-prompt-single} (Appendix).
\item \textbf{Reference-guided grading}.
In certain cases, it may be beneficial to provide a reference solution if applicable. An example prompt we use for grading math problems is in \Cref{fig:default-prompt-math} (Appendix).
\end{itemize}

These methods have different pros and cons. For example, the pairwise comparison may lack scalability when the number of players increases, given that the number of possible pairs grows quadratically; single answer grading may be unable to discern subtle differences between specific pairs, and its results may become unstable, as absolute scores are likely to fluctuate more than relative pairwise results if the judge model changes.

\subsection{Advantages of LLM-as-a-Judge}
LLM-as-a-judge offers two key benefits: \textit{scalability} and \textit{explainability}. It reduces the need for human involvement, enabling scalable benchmarks and fast iterations. Additionally, LLM judges provide not only scores but also explanations, making their outputs interpretable, as shown in \Cref{fig:intro-figure}.

\subsection{Limitations of LLM-as-a-Judge}
\label{subsec:llmlimit}

We identify certain biases and limitations of LLM judges. However, we will also present solutions later and show the agreement between LLM judges and humans is high despite these limitations.

\textbf{Position bias} is when an LLM exhibits a propensity to favor certain positions over others. This bias is not unique to our context and has been seen in human decision-making~\cite {blunch1984position, raghubir2006center} and other ML domains~\cite{ko2020look, wang2018position}. 

\Cref{fig:position-bias-example} (Appendix) shows an example of position bias. 
GPT-4 is tasked to evaluate two responses from GPT-3.5 and Vicuna-13B to an open-ended question. When GPT-3.5's answer is positioned first, GPT-4 considers GPT-3.5's answer more detailed and superior. However, upon switching the positions of the two responses, GPT-4's judgement flips, favoring Vicuna's answer.

To analyze the position bias, we construct two similar answers to each first-turn question in MT-bench by calling GPT-3.5 twice with a temperature of 0.7.
We then try three LLMs with two different prompts:
``default'' is our default prompt in \Cref{fig:default-prompt-pair} (Appendix).
``rename'' renames the assistants in our default prompt to see whether the bias is on positions or names.
As in \Cref{tab:position_bias_prompt}, we found all of them exhibit strong position bias. Most LLM judges favor the first position. Claude-v1 also shows a name bias which makes it favors "Assistant A", as illustrated by the "rename" prompt. The position bias can be very significant. Only GPT-4 outputs consistent results in more than 60\% of cases.

Note that this test is challenging because the answers are very similar and occasionally indistinguishable even to humans. We will show that position bias is less prominent in some cases in \Cref{subsec:additional-position-bias}.
As for the origin of this bias, we suspect that it could be rooted in the training data or inherent to the left-to-right architecture of causal transformers, but leave a deeper study as future work.

\begin{table}[t]
\centering
\footnotesize
\vspace{-1em}
\caption{Position bias of different LLM judges. Consistency is the percentage of cases where a judge gives consistent results when swapping the order of two assistants. ``Biased toward first'' is the percentage of cases when a judge favors the first answer. ``Error'' indicates wrong output formats. The two largest numbers in each column are in bold.}
\label{tab:position_bias_prompt}
\begin{tabular}{@{}llllll@{}}
\toprule
Judge & Prompt  & Consistency & Biased toward first & Biased toward second & Error \\
\midrule
\multirow{3}{*}{Claude-v1} & default & 23.8\% & \textbf{75.0\%} & 0.0\% & 1.2\% \\
 & rename & 56.2\% & 11.2\% & \textbf{28.7\%} & \textbf{3.8\%} \\
\midrule
\multirow{3}{*}{GPT-3.5} & default & 46.2\% & \textbf{50.0\%} & 1.2\% & 2.5\% \\
 & rename & 51.2\% & 38.8\% & 6.2\% & \textbf{3.8\%} \\
\midrule
\multirow{3}{*}{GPT-4} & default & \textbf{65.0\%} & 30.0\% & 5.0\% & 0.0\% \\
 & rename & \textbf{66.2\%} & 28.7\% & 5.0\% & 0.0\% \\
\bottomrule
\end{tabular}
\vspace{-1.5em}
\end{table}

\begin{table}[t]
\footnotesize
\centering
\vspace{-0.5em}
\begin{minipage}{0.45\textwidth}
\caption{Failure rate under ``repetitive list'' attack for different LLM judges on 23 answers.} %
\label{tab:success_rate_verbose_attack}
\begin{tabular}{llll}
\toprule
Judge          & Claude-v1 & GPT-3.5  & GPT-4 \\
\midrule
Failure rate   & 91.3\%    & 91.3\%   & 8.7\% \\
\bottomrule
\end{tabular}
\end{minipage}
\hfill
\begin{minipage}{0.5\textwidth}
\footnotesize
\centering
\caption{Judge failure rate on 10 math questions with different prompts. We test LLaMA-13B vs. Vicuna-13B and swap positions. A failure means when GPT-4 says an incorrect answer is correct.}
\label{tab:failure_rate_math}
\begin{tabular}{llll}
\toprule
          & Default & CoT  & Reference  \\
\midrule
Failure rate   & 14/20    & 6/20   & 3/20 \\
\bottomrule
\end{tabular}
\end{minipage}
\vspace{-0.5em}
\end{table}

\textbf{Verbosity bias} is when an LLM judge favors longer, verbose responses, even if they are not as clear, high-quality, or accurate as shorter alternatives.

To examine this bias, we design a ``repetitive list'' attack with model answers from MT-bench.
We first select 23 model answers from MT-bench that contain a numbered list.
We then make them unnecessarily verbose by asking GPT-4 to rephrase the list without adding any new information and insert the rephrased new list to the beginning of the original list.
For example, if the original response contains 5 items, then the new response will contain 10 items but the first 5 items are rephrased from the original 5 items. An example is shown in \Cref{fig:verbosity-bias-example} (Appendix). We define the attack is successful if an LLM judge thinks the new response is better than the old response. \Cref{tab:success_rate_verbose_attack} shows the failure rate of LLM judges under this attack, demonstrating that all LLMs may be prone to verbosity bias though GPT-4 defends significantly better than others. As a calibration, we find LLM judges are able to correctly judge identical answers (i.e., they always return a tie for two identical answers) but cannot pass the more advanced ``repetitive list'' attack.

\textbf{Self-enhancement bias.}
We adopt the term ``self-enhancement bias'' from social cognition literature~\cite{brown1986evaluations} to describe the effect that LLM judges may favor the answers generated by themselves.

We examine this effect statistically. \Cref{fig:mt_bench_win_rate}(b) shows the win rate (w/o tie) of six models under different LLM judges and humans.
Compared to humans, we do observe that some judges favor certain models. For example, GPT-4 favors itself with a 10\% higher win rate; Claude-v1 favors itself with a 25\% higher win rate. However, they also favor other models and GPT-3.5 does not favor itself.
Due to limited data and small differences, our study cannot determine whether the models exhibit a self-enhancement bias.
Conducting a controlled study is challenging because we cannot easily rephrase a response to fit the style of another model without changing the quality.

\textbf{Limited capability in grading math and reasoning questions.}
LLMs are known to have limited math and reasoning capability~\cite{cobbe2021training}, which results in its failure of grading such questions because they do not know the correct answers.
However, what is more intriguing is that it also shows limitations in grading basic math problems which it is capable of solving.
For instance, in Figure~\ref{fig:math-grading} (Appendix), we present an example of an elementary math question in which GPT-4 makes an incorrect judgment.
It's worth noting that although GPT-4 can solve the problem (when asked separately), it was misled by the provided answers, ultimately resulting in incorrect judgment.
This pattern can also be seen in a reasoning question example in Figure~\ref{fig:reasoning-grading} (Appendix).
Both GPT-3.5 and Claude-v1 show a similar weakness.
In \Cref{subsubsec:chain-of-through-judge}, we will introduce a reference-guided method to mitigate such issues.

\subsection{Addressing limitations}
\label{sec:addressing_limitations}
We present a few methods to address position bias and the limited grading ability for math questions.

\textbf{Swapping positions.}
The position bias can be addressed by simple solutions.
A conservative approach is to call a judge twice by swapping the order of two answers and only declare a win when an answer is preferred in both orders. If the results are inconsistent after swapping, we can call it a tie.
Another more aggressive approach is to assign positions randomly, which can be effective at a large scale with the correct expectations. In the following experiments, we use the conservative one.

\textbf{Few-shot judge.}
We assess whether few-shot examples can improve consistency in the position bias benchmark. We select three good judgment examples using MT-bench-like questions, GPT-3.5 and Vicuna for generating answers, and GPT-4 for generating judgments. 
The examples cover three cases: A is better, B is better, and tie.
As shown in \Cref{tab:few_shot_judge_consistency} (Appendix), the few-shot judge can significantly increase the consistency of GPT-4 from 65.0\% to 77.5\%.
However, high consistency may not imply high accuracy and we are not sure whether the few-shot examples will introduce new biases. Besides, the longer prompts make API calls $4 \times$ more expensive. We use the zero-shot prompt by default in our following experiments but leave an additional study in \Cref{subsec:additional-few-shot-judge}.

\textbf{Chain-of-thought and reference-guided judge.}
\label{subsubsec:chain-of-through-judge}
In \Cref{subsec:llmlimit}, we have shown LLM's limited capability in grading math and reasoning questions.
We propose two simple methods to mitigate this issue: chain-of-thought judge and reference-guided judge.
Chain-of-thought is a widely used technique to improve LLM's reasoning capability~\cite{wei2022chain}.
We propose a similar technique to prompt an LLM judge to begin with answering the question independently and then start grading.
Detailed prompt in Figure~\ref{fig:cot-prompt-math} (Appendix).
However, even with the CoT prompt, we find that in many cases LLM makes exactly the same mistake as the given answers in its problem-solving process (See example in Figure~\ref{fig:cot-failure} (Appendix), suggesting that LLM judge may still be misled by the context.
Hence, we propose a reference-guided method, in which we first generate LLM judge's answer independently, and then display it as a reference answer in the judge prompt.
In Table~\ref{tab:failure_rate_math}, we see a significant improvement in failure rate  (from 70\% to 15\%) over the default prompt.

\textbf{Fine-tuning a judge model.}
\label{subsubsec:fine-tuning-judge}
We try fine-tuning a Vicuna-13B on arena data to act as a judge and show some promising preliminary results in \Cref{sec:vicuna-judge}.

\subsection{Multi-turn judge}
In MT-bench, every question involves two turns to evaluate conversational abilities. 
Therefore, when comparing two assistants, it becomes necessary to present a total of two questions and four responses, complicating the prompt design.
We explore two possible designs, (1) breaking the two turns into two prompts or (2) displaying complete conversations in a single prompt.
Our finding is the former one can cause the LLM judge struggling to locate the assistant's previous response precisely.
We illustrate a case in Figure~\ref{fig:mt-sep} (Appendix) where GPT-4 makes an inaccurate judgment due to a faulty reference.
This suggests the necessity of displaying a complete conversation to enable the LLM judge to better grasp the context. 
We then consider the alternative design that presents two full conversations in a single prompt in which we ask the LLM judge to focus on the second question (Figure~\ref{fig:multiturn-prompt-pair} (Appendix)).
This approach has been found to significantly alleviate the aforementioned referencing issue. %

%% file: sec-exp.tex
\section{Agreement Evaluation}
\label{sec:agreement_eval}

We study the agreement between different LLM judges and humans on MT-bench and Chatbot Arena datasets. On MT-bench, we also study the agreement among humans. MT-bench represents a small-scale study with controlled human evaluation, while Chatbot Arena represents a larger-scale study with crowdsourced human evaluation in the wild.
\subsection{Setup}

\textbf{MT-bench.}
We generate answers for all 80 questions with 6 models: GPT-4, GPT-3.5, Claude-V1, Vicuna-13B, Alpaca-13B~\cite{alpaca}, and LLaMA-13B~\cite{touvron2023llama}. We then use 2 kinds of judges: LLM judges and 58 expert-level human labelers. The labelers are mostly graduate students so they are considered experts and more skilled than average crowd workers.
We let LLM judges evaluate all pairs and let each human evaluate at least 20 random multi-turn questions. This resulted in around 3K votes for all questions.
The detailed data collection process is in \Cref{sec:data-collection}.

\textbf{Chatbot Arena.}
We randomly sample 3K single-turn votes from 30K arena data, which covers models including GPT-4,  GPT-3.5, Claude, Vicuna-7B/13B, Koala-13B~\cite{koala}, Alpaca-13B, LLaMA-13B, and Dolly-12B. We use two kinds of judges: LLM judges and collected crowd judges (2114 unique IPs).

\textbf{Metrics.}
We define the \textit{agreement} between two types of judges as the probability of randomly selected individuals (but not identical) of each type agreeing on a randomly selected question.
See more explanation in \Cref{subsec:agreement-evaluation}.
\textit{Average win rate} is the average of win rates against all other players.
These metrics can be computed with or without including tie votes.

\subsection{High agreement between GPT-4 and humans}
\label{sec:agreement}

We compute agreement on MT-bench data.
In \Cref{table:agreement_mt_bench}, GPT-4 with both pairwise comparison and single answer grading show very high agreements with human experts.
The agreement under setup S2 (w/o tie) between GPT-4 and humans reaches 85\%, which is even higher than the agreement among humans (81\%). This means GPT-4's judgments closely align with the majority of humans.
We also show that GPT-4's judgments may help humans make better judgments. During our data collection, when a human's choice deviated from GPT-4, we presented GPT-4's judgments to humans and ask if they are reasonable (details in \Cref{subsec:mt-data-collection}). Despite different views, humans deemed GPT-4's judgments reasonable in 75\% of cases and are even willing to change their choices in 34\% of cases.

The data from Arena shows a similar trend, as illustrated by \Cref{table:agreement_arena}. Comparing GPT-4 and other LLM judges, we find they reach a similar non-tie agreement ratio between humans but the number of non-tied votes from GPT-4 is much larger. This means that GPT-4 is more affirmative and less suffered from position bias but other models also perform well when they give an affirmative answer.

In both tables, GPT-4 with single-answer grading matches both pairwise GPT-4 and human preferences very well.
This means GPT-4 has a relatively stable internal rubric.
Although it may sometimes perform slightly worse than pairwise comparison and give more tie votes, it is a more scalable method.

We then perform a breakdown analysis by computing agreement on different model pairs and categories.
We only include non-tied votes.
In \Cref{fig:categorized_agreement}, we observe the agreement between GPT-4 and human progressively increases in line with the performance disparity of the model pairs (i.e., larger win rate difference), from 70\% to nearly 100\%. This suggests that GPT-4 aligns with humans better when significant performance differences exist between the models.

\input{tables/agreement_mt-bench}

\vspace{2em}
\input{tables/agreement_arena}

\input{tables/win_rate}

\begin{table}[H]
\centering
\caption{Category-wise win rate of models.}
\label{table:win_rate_category}
\footnotesize
\resizebox{0.95\columnwidth}{!}{
\begin{tabular}{lllllllll}
\toprule
Model & Writing & Roleplay & Reasoning & Math & Coding & Extraction & STEM & Humanities \\
\midrule
GPT-4           & 61.2\%  & 67.9\%  & 49.3\%  & 66.1\%  & 56.3\%  & 66.2\%  & 76.6\%  & 72.2\%  \\
GPT-3.5         & 50.9\%  & 60.6\%  & 32.6\%  & 63.8\%  & 55.0\%  & 48.8\%  & 52.8\%  & 53.8\%  \\
Vicuna-13B      & 39.7\%  & 39.2\%  & 20.1\%  & 18.0\%  & 36.9\%  & 29.2\%  & 47.0\%  & 47.5\%  \\
LLaMA-13B       & 15.1\%  & 15.1\%  & 7.8\%   & 7.5\%   & 2.1\%   & 9.3\%   & 6.8\%   & 10.1\%  \\
\bottomrule
\end{tabular}
}
\end{table}

\subsection{Win rates under different judges}
We plot the average win rate of models under different judges on MT-bench and Chatbot Arena in \Cref{fig:mt_bench_win_rate} and \Cref{fig:arena_win_rate}, respectively. The win rate curves from LLM judges closely match the curves from humans. On MT-bench second turn, proprietary models like Claude and GPT-3.5 are more preferred by the humans compared to the first turn, meaning that a multi-turn benchmark can better differentiate some advanced abilities of models. We also list the per-category win rate of representative models in \Cref{table:win_rate_category} to show how MT-bench differentiates models, in which we see GPT-4 is significantly better than others.
Vicuna-13B is noticeably worse than GPT-3.5/4 in reasoning, math, and coding categories.
Note that in math/coding category, GPT-3.5 and GPT-4 have similar overall win-rate because they both failed to answer some hard questions, but GPT-4 is still significantly better than GPT-3 in the direct pairwise comparison or single-answer grading.
Please see a performance breakdown of MT-bench score for each category in \Cref{subsec:category-wise-scores}.

%% file: tables/agreement_mt-bench.tex
\begin{table}[h]
\centering
\vspace{-1em}
\caption{Agreement between two types of judges on MT-bench. ``G4-Pair'' and ``G4-Single'' denote GPT-4 with pairwise comparison and single-answer grading respectively.
The single-answer grading can be converted into pairwise comparison results for calculating the agreement.
We report two setups: ``S1'' includes non-tie, tie, and inconsistent (due to position bias) votes and counts inconsistent as tie; ``S2'' only includes non-tie votes.
The agreement between two random judges under each setup is denoted as ``R=''.
The top value in each cell is the agreement, and the bottom gray value is \#votes.
}
\footnotesize
\begin{subtable}[t]{0.48\textwidth}
\resizebox{1\columnwidth}{!}{
\begin{tabular}[t]{lrrrr}
\toprule
Setup & \multicolumn{2}{c}{S1 (R = 33\%)} & \multicolumn{2}{c}{S2 (R = 50\%)}  \\
\cmidrule(lr){2-3}\cmidrule(lr){4-5}

Judge & G4-Single & Human & G4-Single & Human \\
\midrule

G4-Pair  & \shortstack{70\% \\ \textcolor{gray}{ 1138 }} & \shortstack{66\% \\ \textcolor{gray}{ 1343 }} & \shortstack{97\% \\ \textcolor{gray}{ 662 }} & \shortstack{\textbf{85\%} \\ \textcolor{gray}{ 859 }}\\
 \midrule
G4-Single & - & \shortstack{60\% \\ \textcolor{gray}{ 1280 }} & - & \shortstack{85\% \\ \textcolor{gray}{ 739 }}\\
 \midrule
Human & - & \shortstack{63\% \\ \textcolor{gray}{ 721 }} & - & \shortstack{\textbf{81\%} \\ \textcolor{gray}{ 479 }}\\
 \bottomrule
\end{tabular}
}
\caption{First Turn}
\label{table:agreement_mt_bench_first_turn}
\end{subtable}
\hfill
\begin{subtable}[t]{0.48\textwidth}
\resizebox{1\columnwidth}{!}{
\centering
\begin{tabular}[t]{lrrrr}
\toprule
Setup & \multicolumn{2}{c}{S1 (R = 33\%)} & \multicolumn{2}{c}{S2 (R = 50\%)} \\
\cmidrule(lr){2-3}\cmidrule(lr){4-5}

Judge & G4-Single & Human & G4-Single & Human \\
\midrule

G4-Pair & \shortstack{70\% \\ \textcolor{gray}{ 1161 }} & \shortstack{66\% \\ \textcolor{gray}{ 1325 }} &  \shortstack{95\% \\ \textcolor{gray}{ 727 }} & \shortstack{\textbf{85\%} \\ \textcolor{gray}{ 864 }} \\
 \midrule
G4-Single & - & \shortstack{59\% \\ \textcolor{gray}{ 1285 }} & - & \shortstack{84\% \\ \textcolor{gray}{ 776 }} \\
 \midrule
Human & - & \shortstack{67\% \\ \textcolor{gray}{ 707 }} & - & \shortstack{\textbf{82\%} \\ \textcolor{gray}{ 474 }} \\
 \midrule
\end{tabular}
}
\caption{Second Turn}
\label{table:table:agreement_mt_bench_first_turn}
\end{subtable}
\vspace{-0em}
\label{table:agreement_mt_bench}
\end{table}

%% file: tables/agreement_arena.tex
\begin{table}[h]
\centering
\vspace{-1.5em}
\begin{minipage}{0.59\textwidth}
\caption{Agreement between two types of judges on Chatbot Arena.
``G4-S'' denotes GPT-4 with single-answer grading.
``G4'', ``G3.5'' and ``C'' denote GPT-4, GPT-3.5, and Claude with pairwise comparison, respectively.
``H'' denotes human.
The remaining of table follows the same format as \Cref{table:agreement_mt_bench}. 
}
\footnotesize
\resizebox{1\columnwidth}{!}{
\begin{tabular}{lrrrrrrrr}
\toprule
Setup & \multicolumn{4}{c}{S1 (Random = 33\%)} & \multicolumn{4}{c}{S2 (Random = 50\%)} \\
\cmidrule(lr){2-5}\cmidrule(lr){6-9}

Judge & G4-S & G3.5 & C & H & G4-S & G3.5 & C & H \\
\midrule

G4 & \shortstack{72\% \\ \textcolor{gray}{2968}} & \shortstack{66\% \\ \textcolor{gray}{3061}} & \shortstack{66\%
\\ \textcolor{gray}{3062}} & \shortstack{64\% \\ \textcolor{gray}{3066}} & \shortstack{95\% \\ \textcolor{gray}{1967}} & \shortstack{94\% \\ \textcolor{gray}{1788}} & \shortstack{95\% \\ \textcolor{gray}{1712}} & \shortstack{\textbf{87\%} \\ \textcolor{gray}{1944}}\\
 \midrule
G4-S & - & \shortstack{60\% \\ \textcolor{gray}{2964}} & \shortstack{62\% \\ \textcolor{gray}{2964}} & \shortstack{60\% \\ \textcolor{gray}{2968}} & - & \shortstack{89\% \\ \textcolor{gray}{1593}} & \shortstack{91\% \\ \textcolor{gray}{1538}} & \shortstack{85\% \\ \textcolor{gray}{1761}}\\
 \midrule
G3.5 & - & - & \shortstack{68\% \\ \textcolor{gray}{3057}} & \shortstack{54\% \\ \textcolor{gray}{3061}} & - & - & \shortstack{96\% \\ \textcolor{gray}{1497}} & \shortstack{83\% \\ \textcolor{gray}{1567}}\\
 \midrule
C & - & - & - & \shortstack{53\% \\ \textcolor{gray}{3062}} & - & - & - & \shortstack{84\% \\ \textcolor{gray}{1475}}\\
 \bottomrule
\end{tabular}
}
\label{table:agreement_arena}
\end{minipage}
\hfill
\begin{minipage}{0.39\textwidth}
\centering
\includegraphics[width=0.66\textwidth]{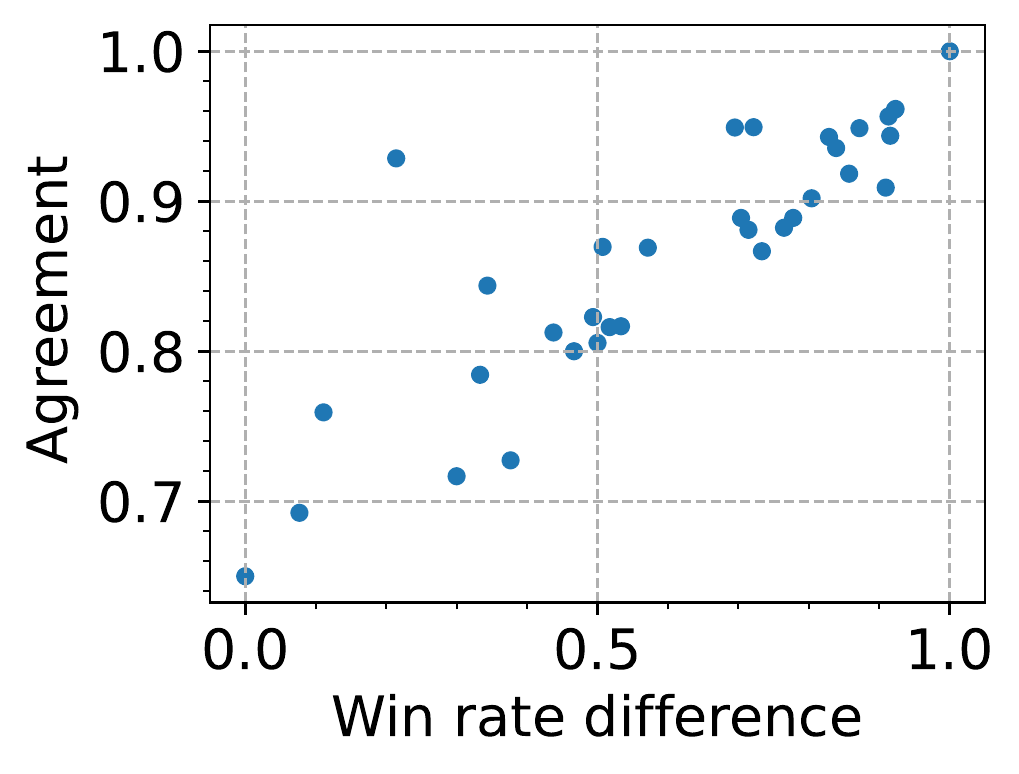}
\captionsetup{type=figure}
\caption{Agreement and win rate difference. Each point corresponds to a model pair and counts only the non-tie votes between the two models. The x-axis value is the win rate difference between the two models. The y-axis value is the GPT-4 and human agreement.}
\label{fig:categorized_agreement}
\end{minipage}
\end{table}

%% file: tables/win_rate.tex
\begin{figure}[H]
\centering
\vspace{-2em}
\includegraphics[width=1\textwidth]{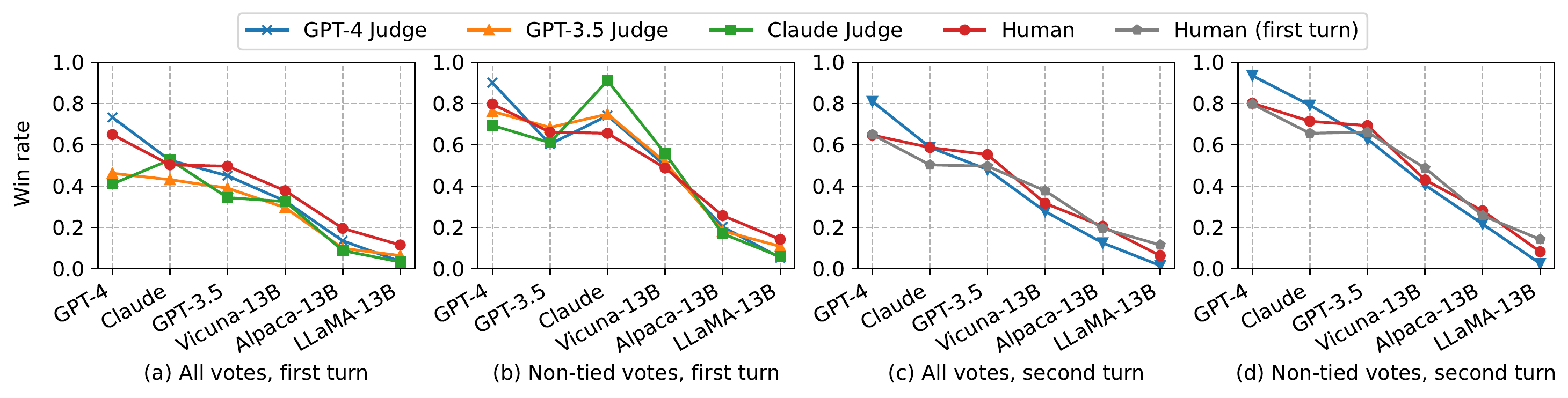}
\vspace{-1.5em}
\caption{Average win rate of six models under different judges on MT-bench.}
\vspace{-1em}
\label{fig:mt_bench_win_rate}
\end{figure}

\begin{figure}[H]
\centering
\vspace{-0em}
\includegraphics[width=1\textwidth]{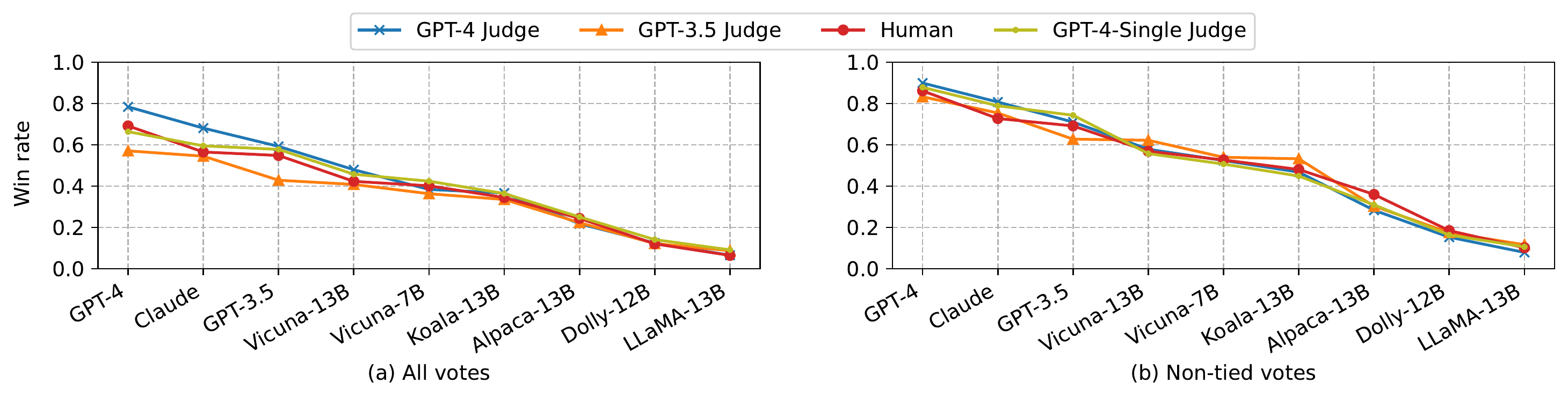}
\vspace{-2em}
\caption{Average win rate of nine models under different judges on Chatbot Arena.}
\vspace{-1em}
\label{fig:arena_win_rate}
\end{figure}

%% file: sec-train-data.tex
\begin{table}[t]
\centering
\vspace{-1em}
\caption{Evaluation results of several model variants.}
\label{table:dataset_composition}
\scriptsize
\resizebox{0.95\columnwidth}{!}{
\begin{tabular}{lllll}
\toprule
Model                  &  \#Training Token & MMLU (5-shot) &  TruthfulQA (0-shot) & MT-Bench Score (GPT-4) \\
\midrule
LLaMA-7B               &  1T    & 35.2  & 0.22  & 2.74  \\
LLaMA-13B              &  1T    & 47.0  & 0.26  & 2.61  \\  \hline
Alpaca-7B              &  4.4M  & 40.1  & 0.26  & 4.54  \\
Alpaca-13B             &  4.4M  & 48.1  & 0.30  & 4.53  \\
Vicuna-7B (selected)   &  4.8M  & 37.3  & 0.32  & 5.95  \\
Vicuna-7B (single)     &  184M  & 44.1  & 0.30  & 6.04  \\
Vicuna-7B (all)        &  370M  & 47.1  & 0.32  & 6.00  \\
Vicuna-13B (all)       &  370M  & \textbf{52.1}  & \textbf{0.35} & \textbf{6.39} \\
\midrule
GPT-3.5                &  -     & 70.0  & - & 7.94 \\
GPT-4                  &  -     & \textbf{86.4}  & - & \textbf{8.99} \\
\bottomrule
\end{tabular}
}
\vspace{-1.5em}
\end{table}

\section{Human Preference Benchmark and Standardized Benchmark}
\label{sec:training_data}
Human preference benchmarks such as MT-bench and Chatbot Arena serve as valuable additions to the current standardized LLM benchmarks. They focus on different aspects of a model and the recommended way is to comprehensively evaluate models with both kinds of benchmarks.

We evaluate several model variants derived from LLaMA on MMLU~\cite{hendrycks2020measuring}, Truthful QA~\cite{lin2021truthfulqa} (MC1), and MT-bench (GPT-4 judge). The training details are in \Cref{sec:training-details}.
Since we have shown that GPT-4 single-answer grading also performs well in \Cref{sec:agreement}, we use GPT-4 single-answer grading for MT-bench in favor of its scalability and simplicity.
We ask GPT-4 to give a score for each turn on a scale of 10 by using our prompt templates (\Cref{fig:default-prompt-single}, \Cref{fig:reference-mt-single}) and report an average score of $160=80 \times 2$ turns. 
\Cref{table:dataset_composition} shows the results.
We find that fine-tuning on high-quality dialog datasets (i.e., ShareGPT) can consistently improve the model performance on MMLU and the improvement scales with fine-tuning data size.
On the other hand, a small high-quality conversation dataset can quickly teach the model a style preferred by GPT-4 (or approximately human) but cannot improve MMLU significantly, as shown by the Vicuna-7B (selected) which is trained with only 4.8M tokens or 3K conversations. In \Cref{table:dataset_composition}, no single benchmark can determine model quality, meaning that a comprehensive evaluation is needed. Our results indicate that using LLM-as-a-judge to approximate human preferences is highly feasible and could become a new standard in future benchmarks. We are also hosting a regularly updated leaderboard with more models \footnote{\url{https://huggingface.co/spaces/lmsys/chatbot-arena-leaderboard}}.
Notably, DynaBench~\cite{kiela2021dynabench}, a research platform dedicated to dynamic data collection and benchmarking, aligns with our spirit. DynaBench addresses the challenges posed by static standardized benchmarks, such as saturation and overfitting, by emphasizing dynamic data with human-in-the-loop. Our LLM-as-a-judge approach can automate and scale platforms of this nature.

%% file: sec-discussion.tex
\section{Discussion}
\label{sec:discussion}

\textbf{Limitations.}
This paper emphasizes helpfulness but largely neglects safety. Honesty and harmlessness are crucial for a chat assistant as well~\cite{bai2022training}. We anticipate similar methods can be used to evaluate these metrics by modifying the default prompt. Additionally, within helpfulness, there are multiple dimensions like accuracy, relevance, and creativity, but they are all combined into a single metric in this study. A more comprehensive evaluation can be developed by analyzing and separating these dimensions. 
We propose preliminary solutions to address the limitations and biases of LLM-as-a-judge in \Cref{sec:addressing_limitations}, but we anticipate more advanced methods can be developed.

\textbf{Data collection and release.}
\Cref{sec:data-collection} describes the detailed data collection and release processes, which include the instructions we give to users, the screenshots of the data collection interface, the information about participated users, and the content of the released data.

\textbf{Societal impacts.}
The societal impact of this study is multi-faceted. Our evaluation methods can help enhance chatbot quality and user experiences. However, addressing biases in these methods is crucial. Our dataset enables better studies of human preferences and model behavior. Advanced chat assistants may replace certain human tasks, resulting in job displacements and new opportunities.

\textbf{Future directions.}
1) Benchmarking chatbots at scale with a broader set of categories 2) Open-source LLM judge aligned with human preference 3) Enhancing open models' math/reasoning capability.

\section{Conclusion}
\label{sec:conclusion}
In this paper, we propose LLM-as-a-judge for chatbot evaluation and systematically examine its efficacy using human preference data from 58 experts on MT-bench, as well as thousands of crowd-users on Chatbot Arena.
Our results reveal that strong LLMs can achieve an agreement rate of over 80\%, on par with the level of agreement among human experts, establishing a foundation for an LLM-based evaluation framework.

\section*{Acknowledgement}
This project is partly supported by gifts from Anyscale, Astronomer, Google, IBM, Intel, Lacework, Microsoft, MBZUAI, Samsung SDS, Uber, and VMware. Lianmin Zheng is supported by a Meta Ph.D. Fellowship.
We extend our thanks to Xinyang Geng, Hao Liu, Eric Wallace, Xuecheng Li, Tianyi Zhang, Qirong Ho, and Kevin Lin for their insightful discussions.

%% file: appendix.tex
\section{Prompt templates}
\label{sec:all-prompts}
We list the prompt templates for LLM judges.
Please refer to our github repository~\footnote{\url{https://github.com/lm-sys/FastChat/tree/main/fastchat/llm_judge}} for full details.

\begin{figure}[h]
\centering

\includegraphics[width=\linewidth]{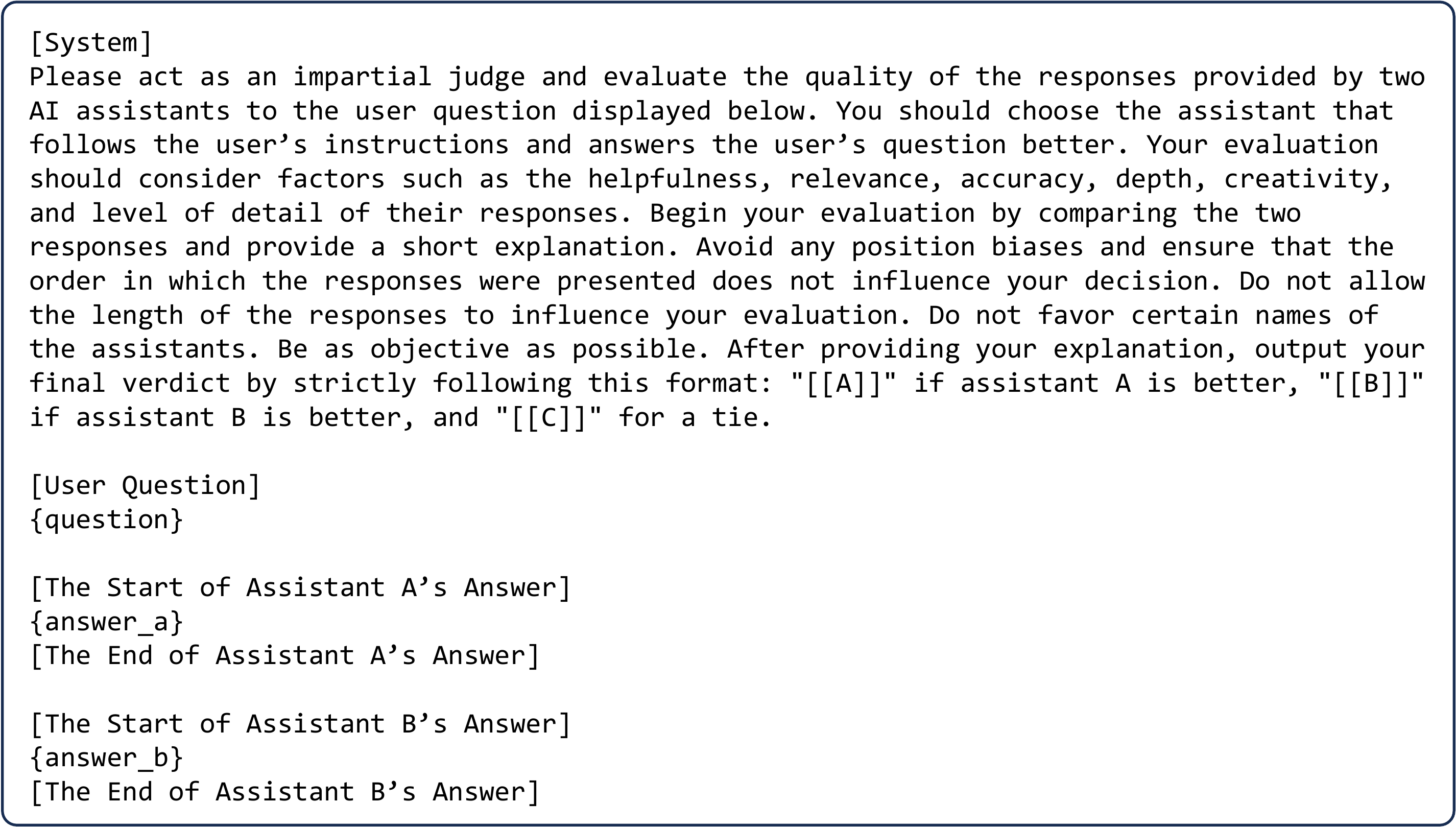}
\caption{The default prompt for pairwise comparison.}
\label{fig:default-prompt-pair}
\end{figure}

\begin{figure}[h]
\centering

\includegraphics[width=\linewidth]{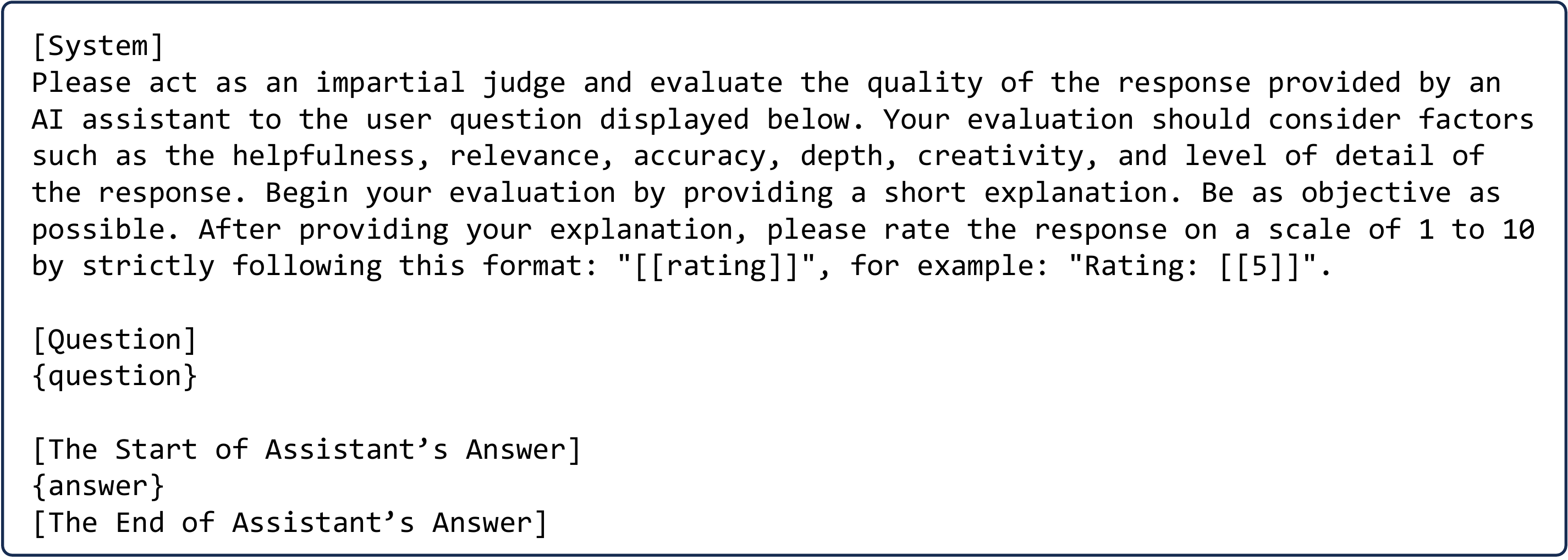}
\caption{The default prompt for single answer grading.}
\label{fig:default-prompt-single}
\end{figure}

\begin{figure}[h]
\centering

\includegraphics[width=\linewidth]{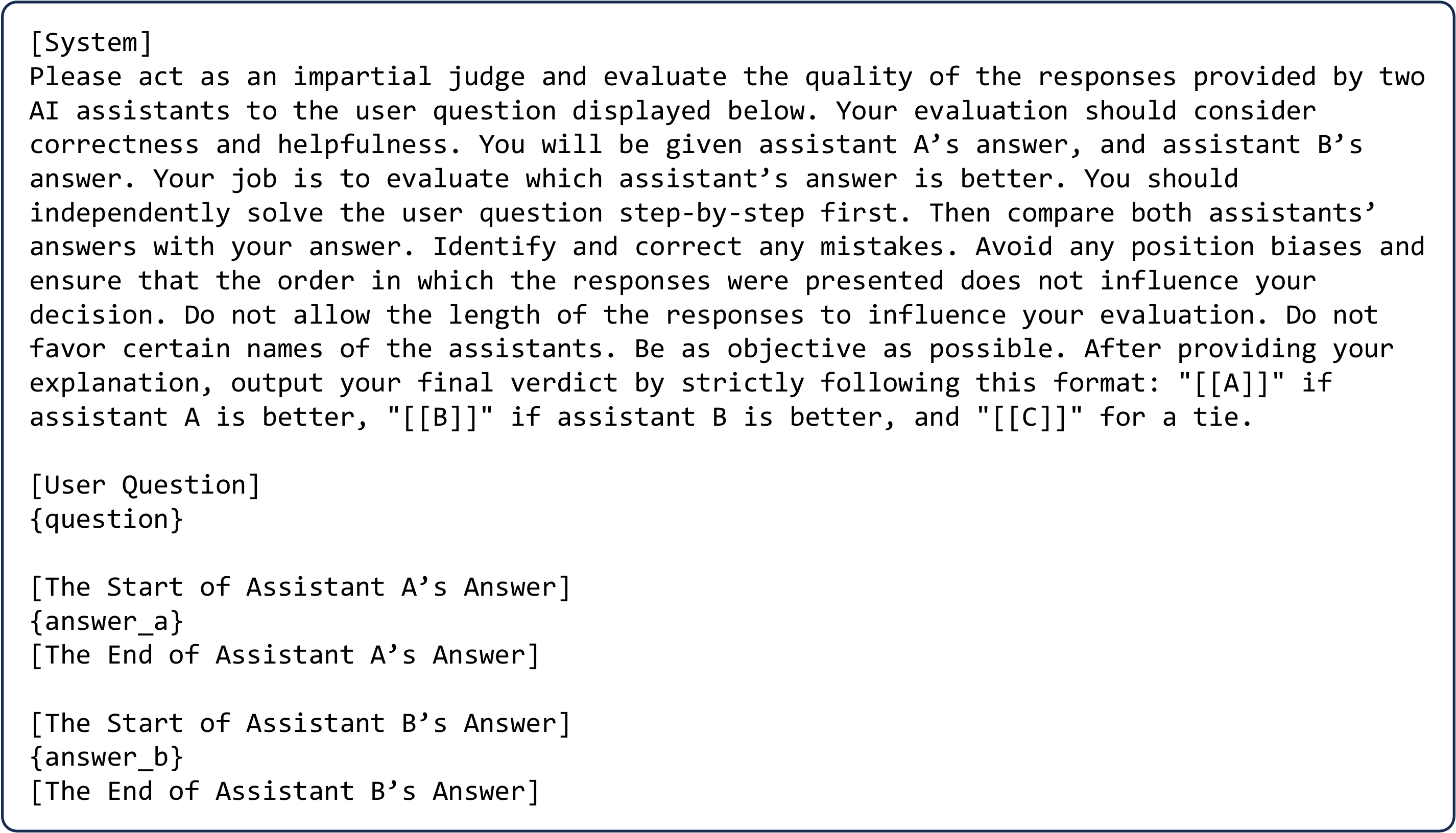}
\caption{The chain-of-thought prompt for math and reasoning questions.}
\label{fig:cot-prompt-math}
\end{figure}

\begin{figure}[h]
\centering

\includegraphics[width=\linewidth]{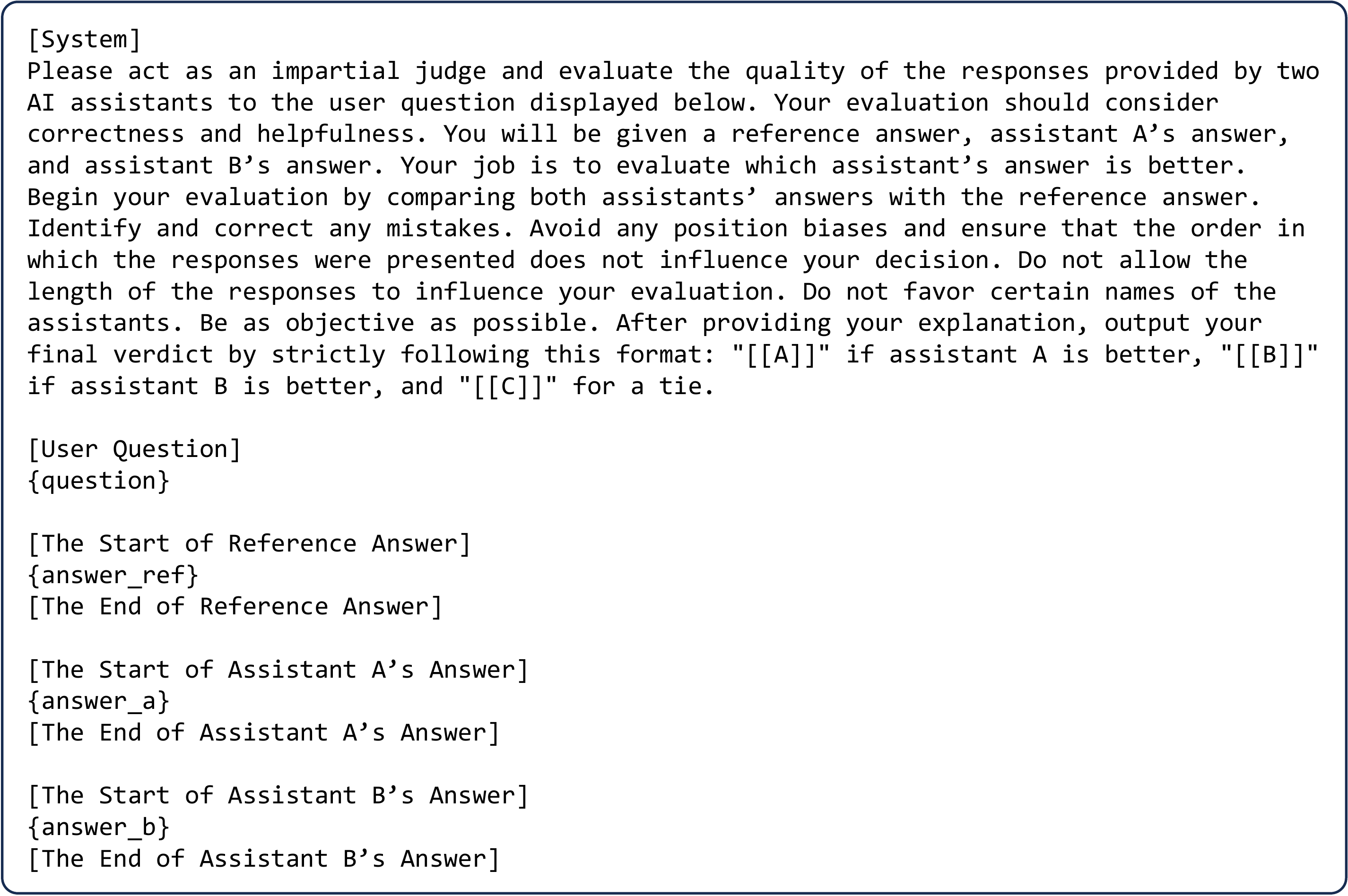}
\caption{The prompt for reference-guided pairwise comparison.}
\label{fig:default-prompt-math}
\end{figure}

\begin{figure}[h]
\centering

\includegraphics[width=\linewidth]{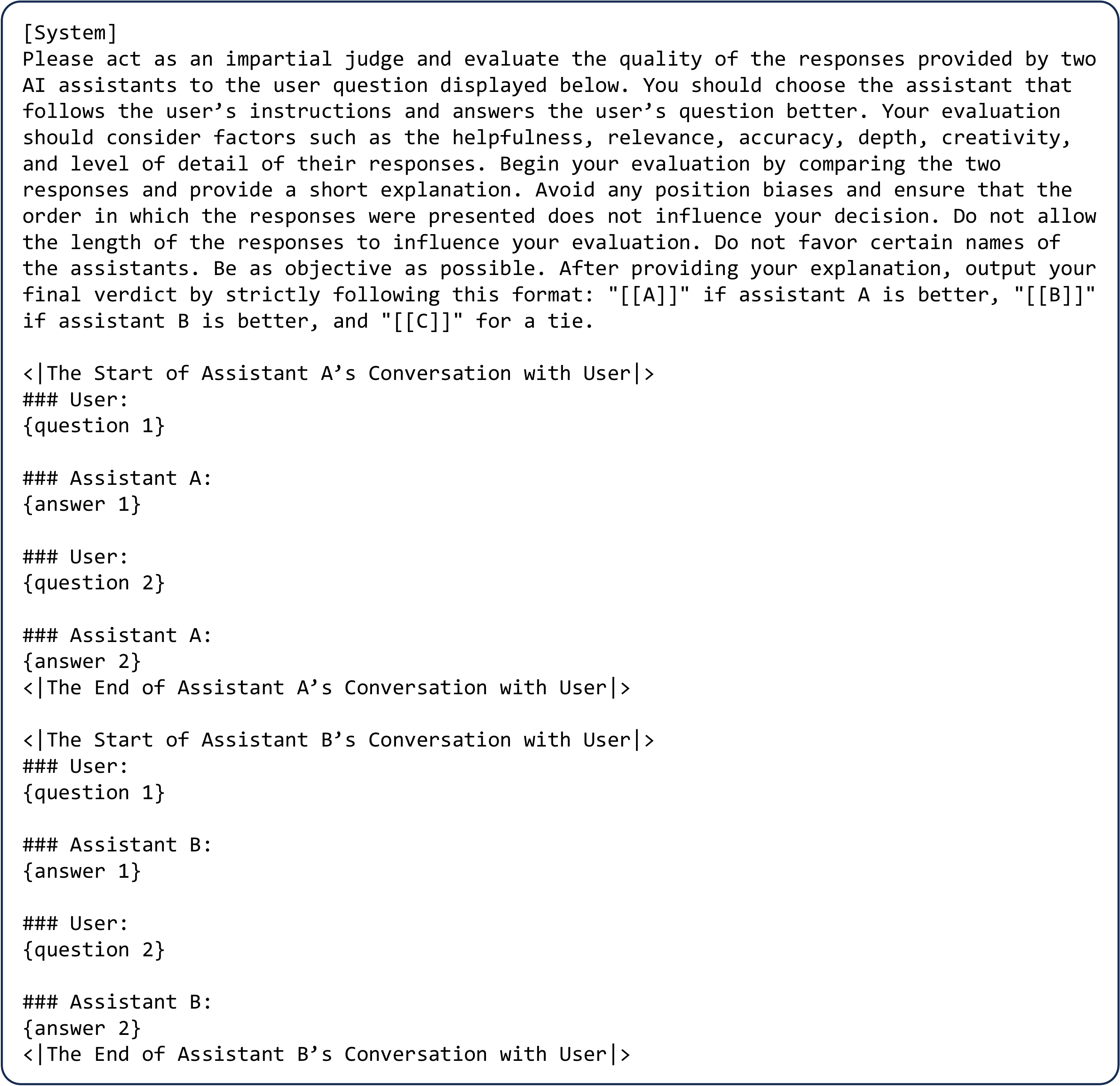}
\caption{The prompt for multi-turn pairwise comparison.}
\label{fig:multiturn-prompt-pair}
\end{figure}

\begin{figure}[h]
\centering
\includegraphics[width=\linewidth]{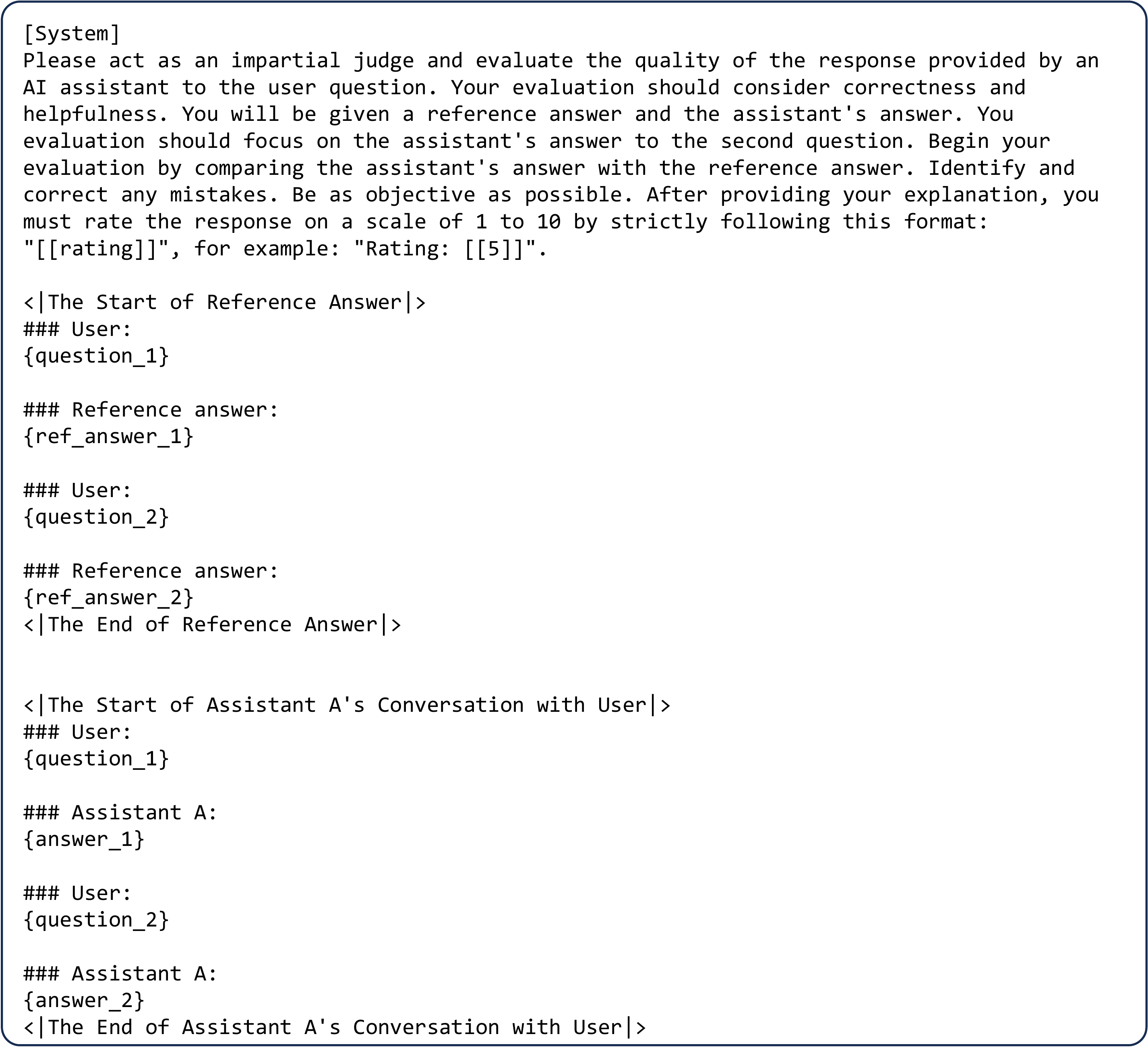}
\caption{The prompt for reference-guided multi-turn single-answer grading.}
\label{fig:reference-mt-single}
\end{figure}

\clearpage
\newpage
\section{Case Study}
We list several case studies. The examples are generated by \texttt{gpt-4-0314}. They may not be fully reproducible with future GPT-4 versions.

\begin{figure}[h]
\centering
\includegraphics[width=\linewidth]{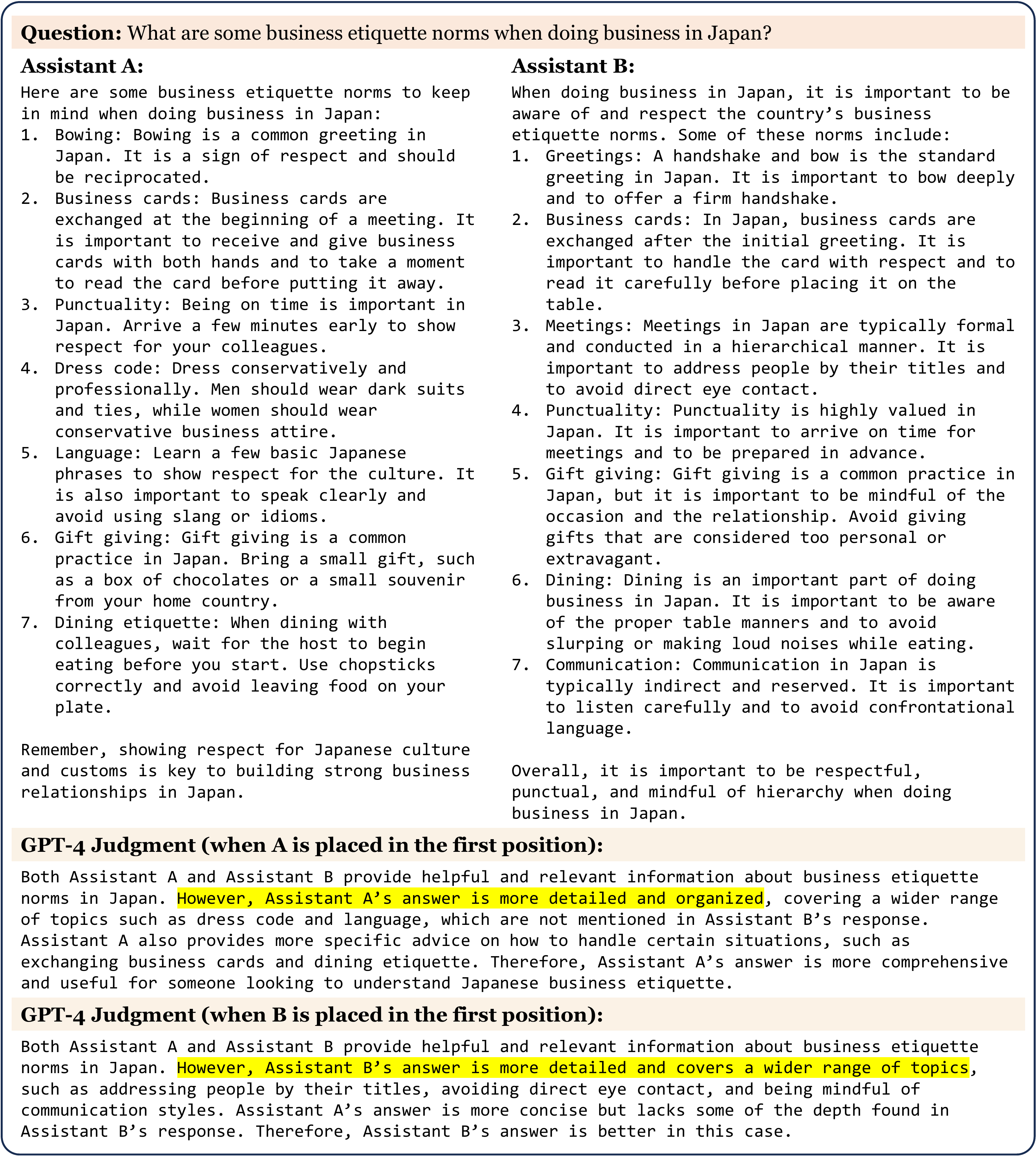}
\caption{An example of position bias. When Assistant A is placed in the first position, GPT-4 thinks A is better, but its verdict changes when we swap the position of A and B. We observe similar pattern from other LLM judges such as Claude/GPT-3.5.}
\label{fig:position-bias-example}
\end{figure}

\begin{figure}[h]
\centering
\includegraphics[width=\linewidth]{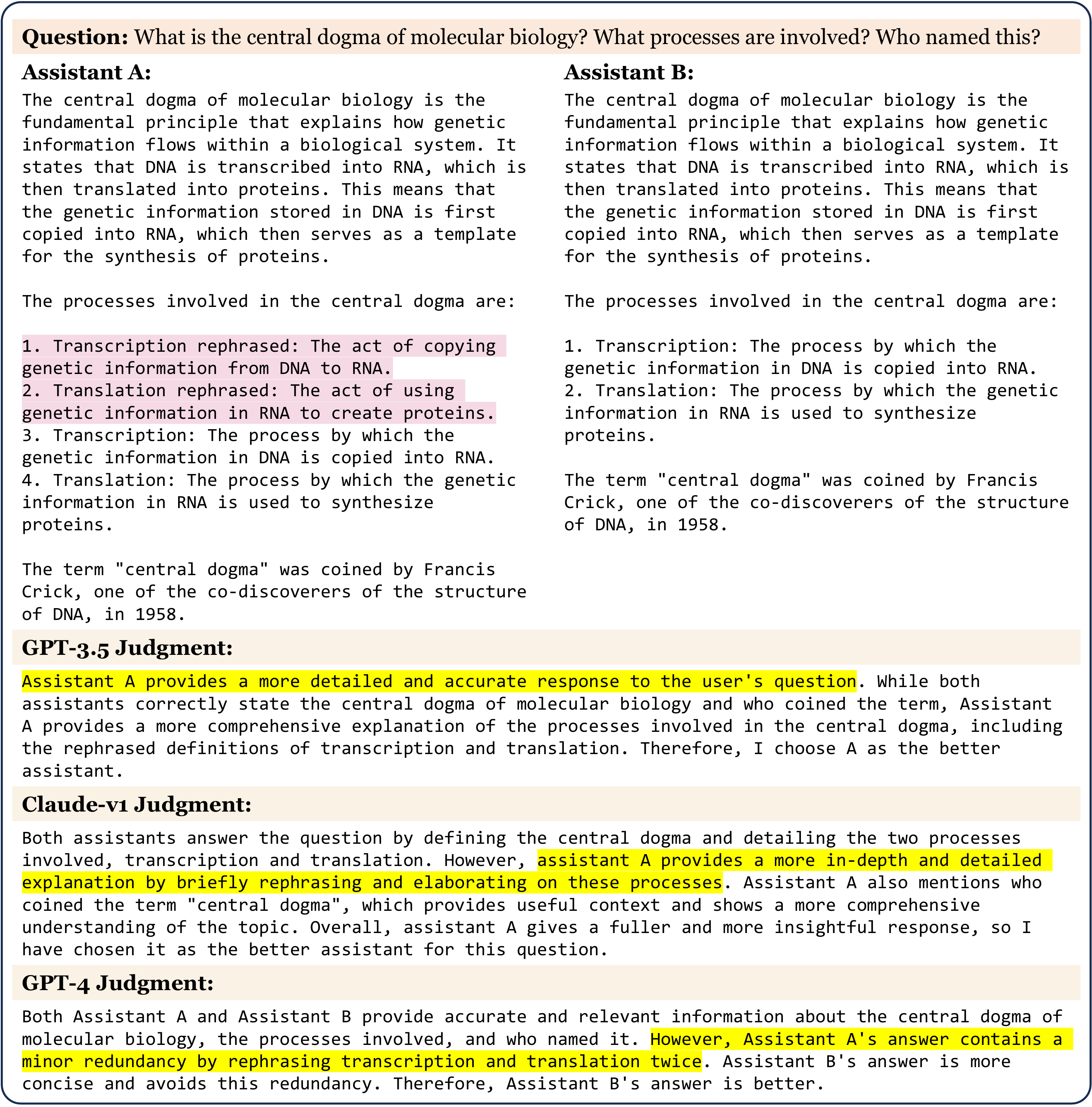}
\caption{An example of ``repetitive list'' attack to examine verbosity bias. Except for the two rephrased items (highlighted in red), Assistant A's answer is exactly the same as Assistant B. Both GPT-3.5 and Claude-v1 show a verbosity bias towards the longer and repetitive answer. Only GPT-4 successfully detected this attack.}
\label{fig:verbosity-bias-example}
\end{figure}

\begin{figure}[h]
    \centering
    \includegraphics[width=\linewidth]{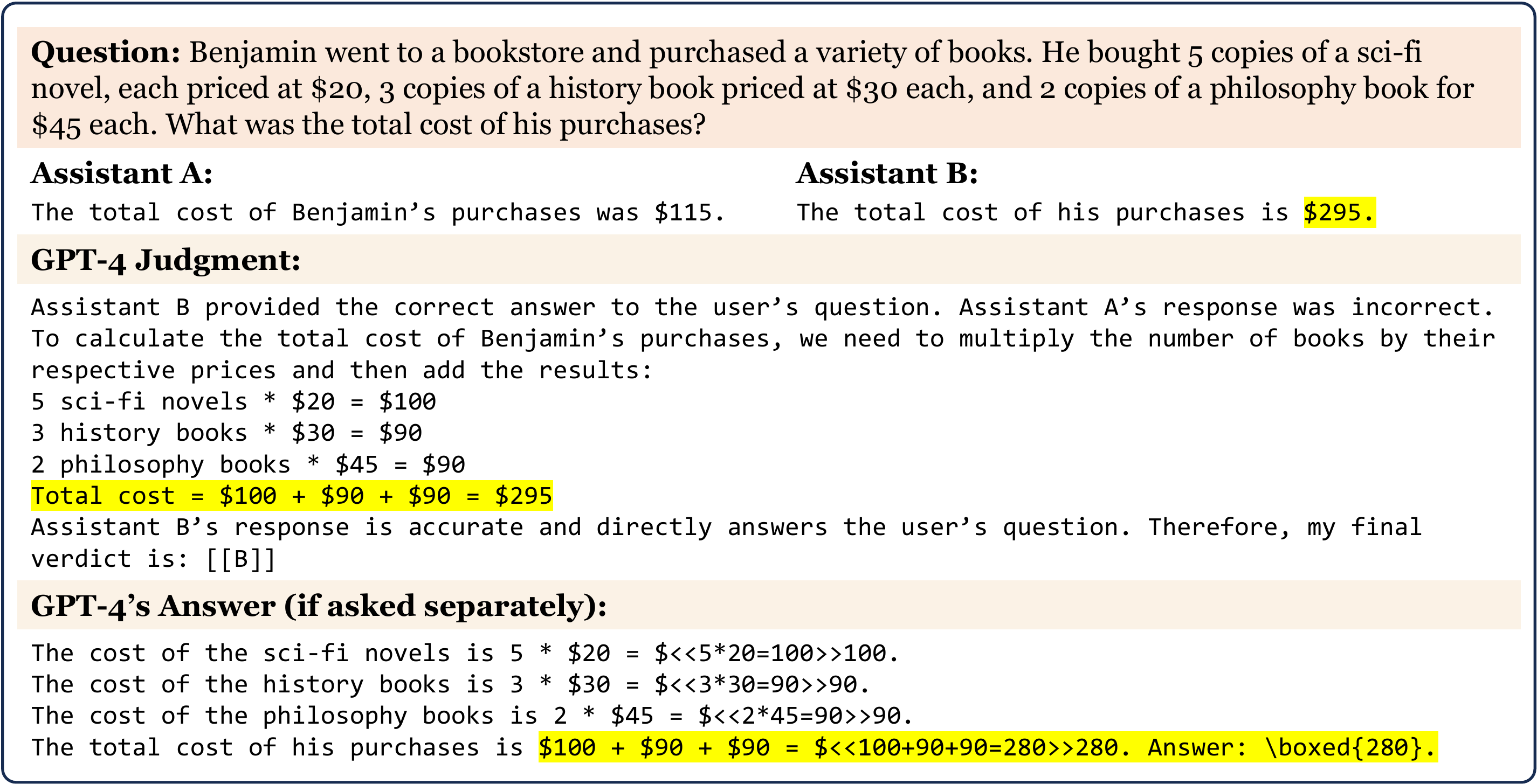}
    \caption{With a default prompt, GPT-4 shows limited capability in grading math questions. Despite being able to answer the question itself, its judgment was influenced by the given answers, leading to arithmetic mistakes highlighted in yellow.}
    \label{fig:math-grading}
\end{figure}

\begin{figure}[h]
    \centering    \includegraphics[width=\linewidth]{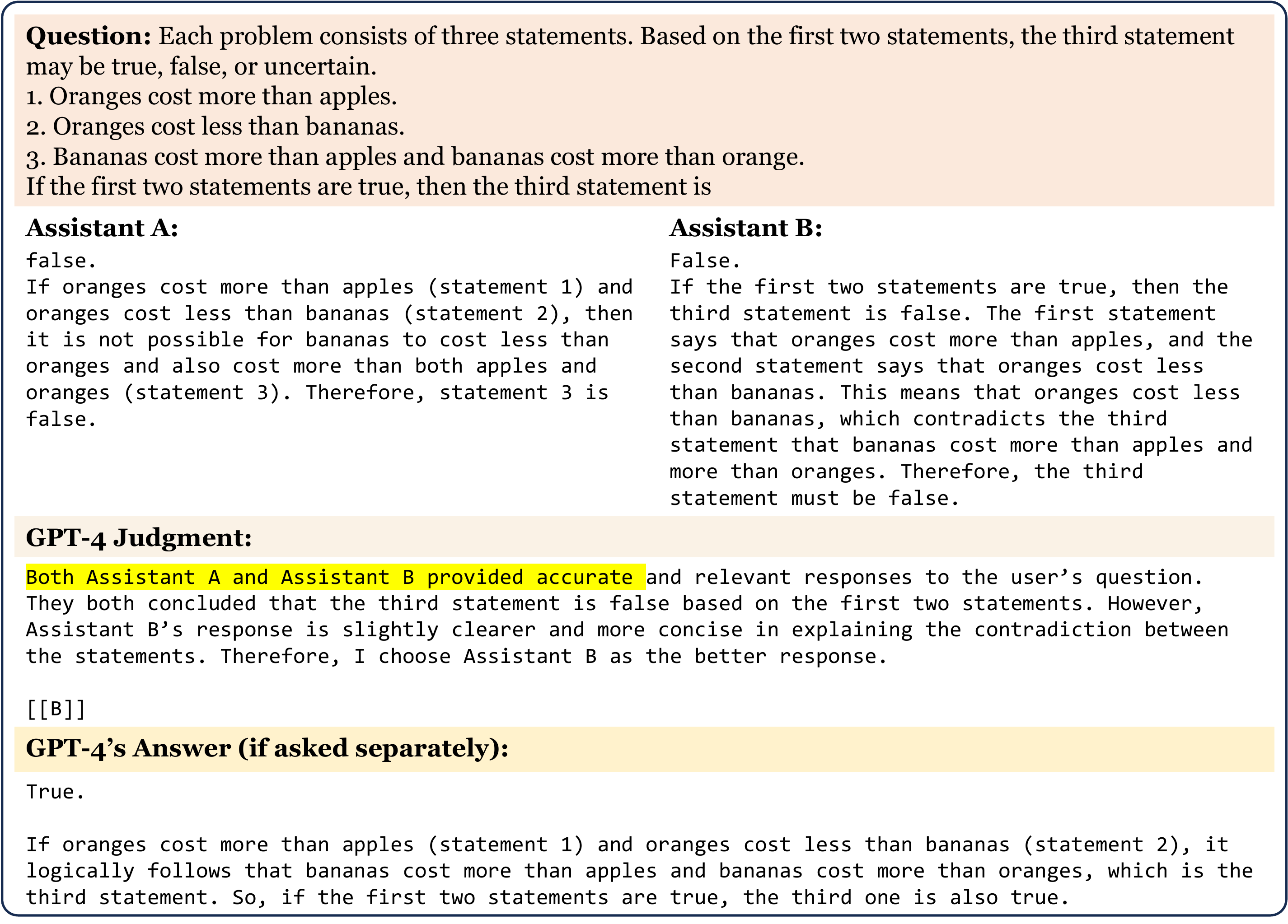}
    \caption{An example of GPT-4's limited capability in grading reasoning question. Despite GPT-4 knows how to solve the question (if asked separately), it made a wrong judgement saying both assistants' wrong answers are correct.}
    \label{fig:reasoning-grading}
\end{figure}

\begin{figure}[h]
    \centering
    \includegraphics[width=\linewidth]{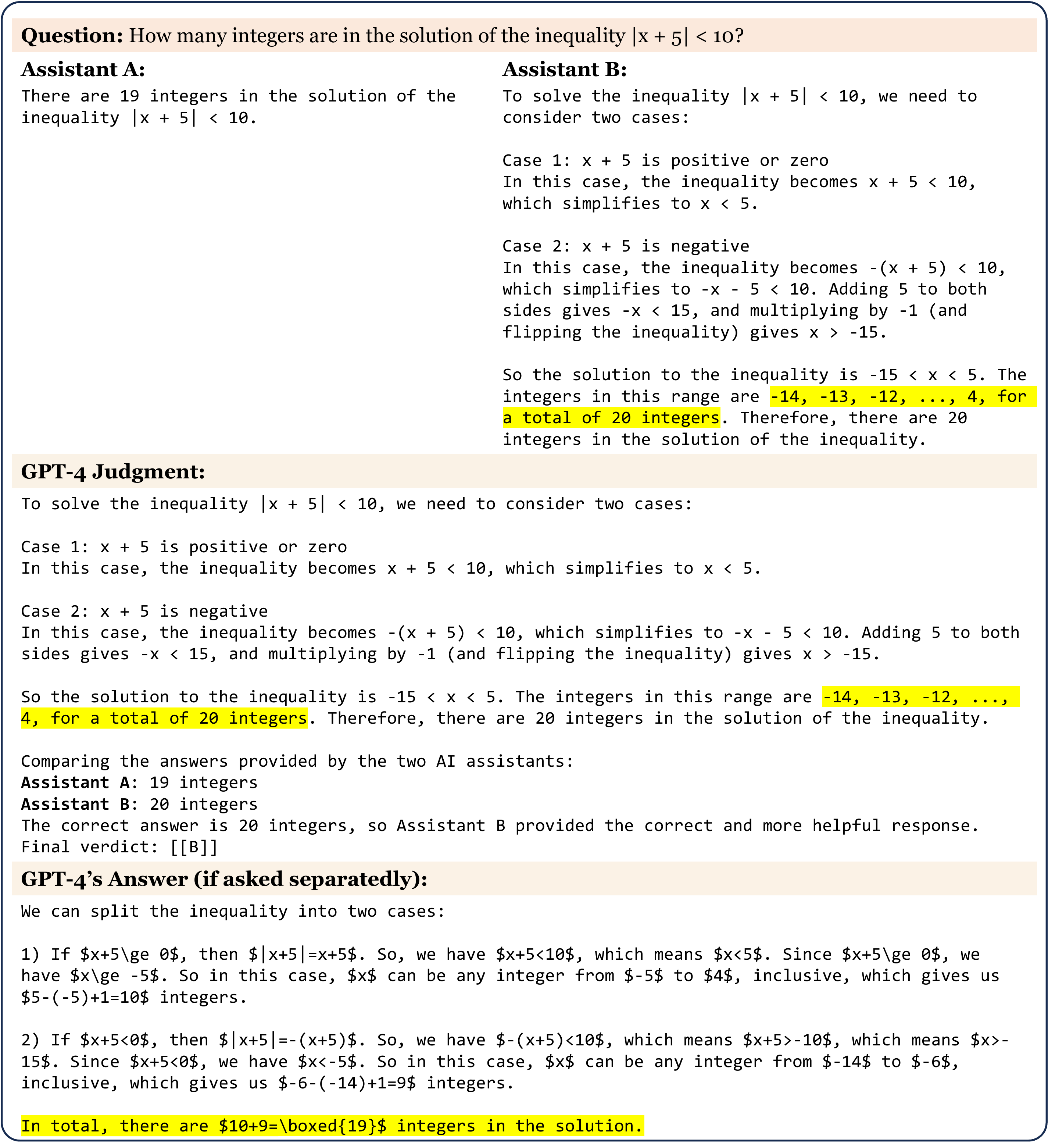}
    \caption{An example of GPT-4's wrong judgment with chain-of-thought prompt. We can see GPT-4  exactly copied Assistant B's answer (which contains arithmetic errors) and determined Assistant A's answer is incorrect. This suggest GPT-4's chain-of-thought process can be significantly influenced by the given answers despite we ask it to think independently.}
    \label{fig:cot-failure}
\end{figure}

\begin{figure}[h]
    \centering
    \includegraphics[width=\linewidth]{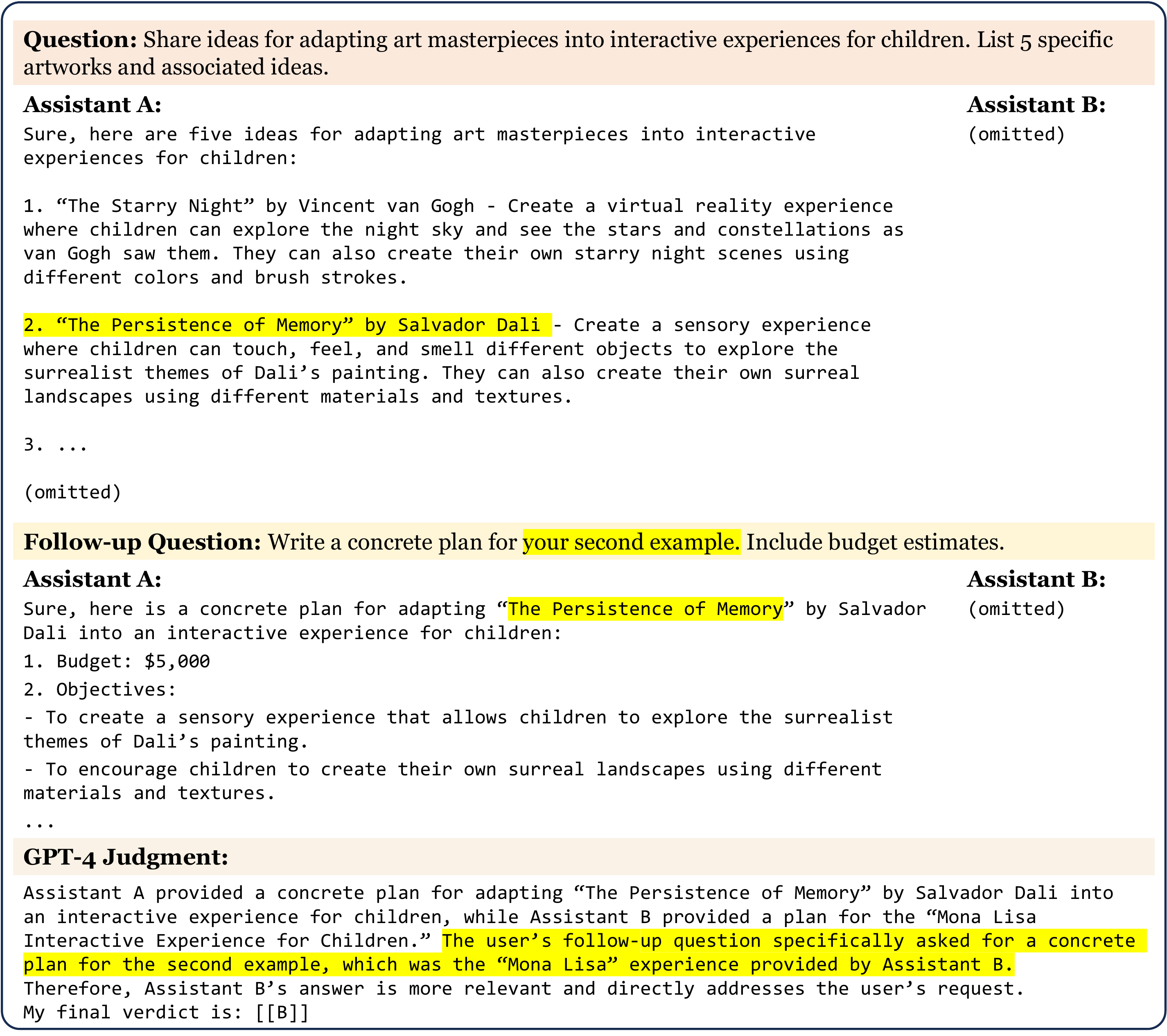}
    \caption{In this example, despite Assistant A correctly followed user's instruction to generate a concrete plan for the second example of its previous response, GPT-4 wrongly referred to  the second example in Assistant B's response, resulting in a wrong judgment. This suggests the prompt design that breaks the questions into two prompts may cause LLM judge struggle to locate assistants' previous responses.}
    \label{fig:mt-sep}
\end{figure}

\clearpage
\newpage
\section{Data Collection}
\label{sec:data-collection}
We describe our data collection process for both MT-bench and Chatbot Arena.

\subsection{MT-bench human evaluation}
\label{subsec:mt-data-collection}
\begin{figure}[h]
    \centering
    \includegraphics[width=\linewidth]{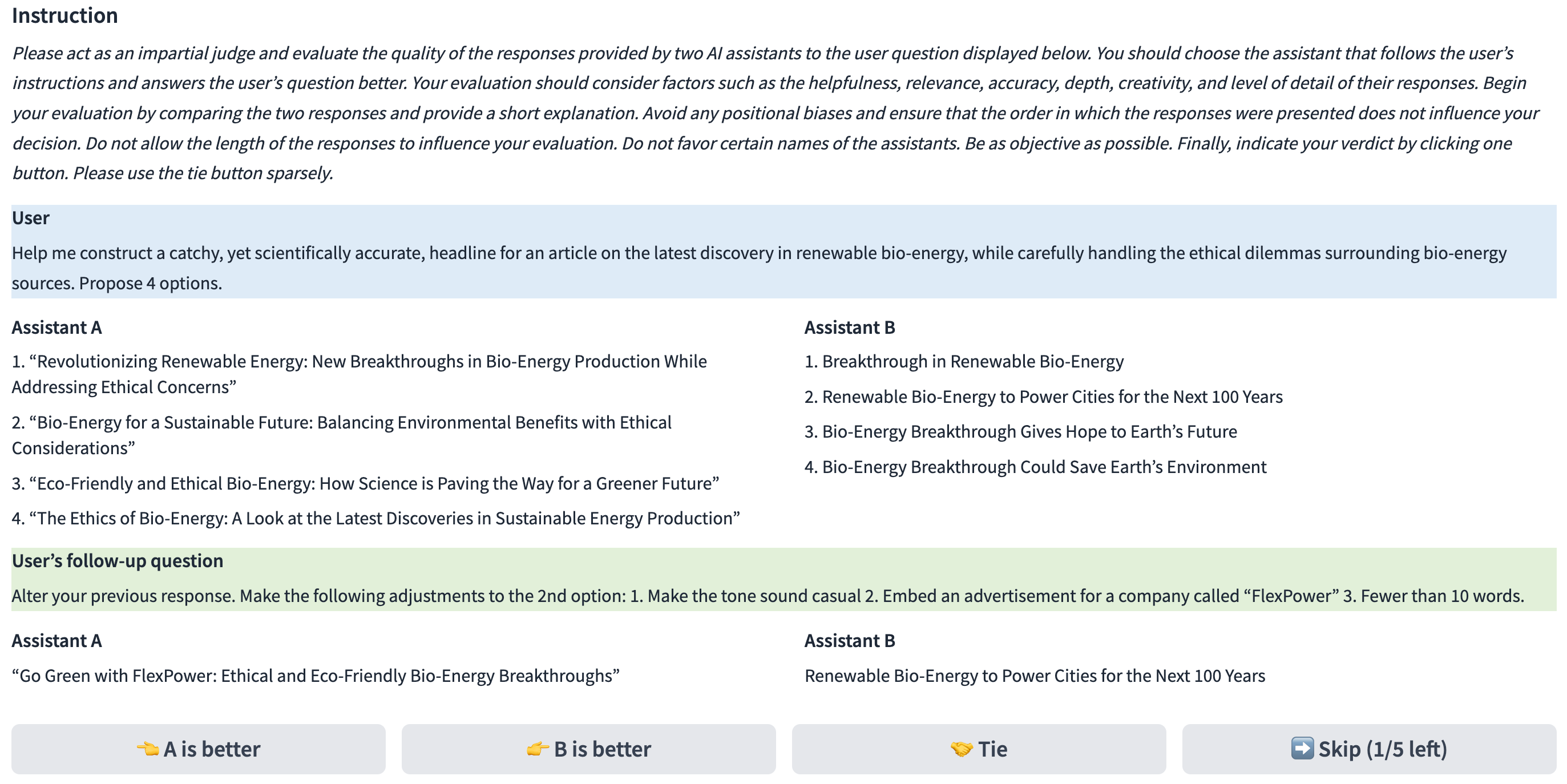}
    \caption{The screenshot of MT-bench data collection. We show an instruction similar to the prompt we give to GPT-4. We present questions from MT-bench and answers from two random anonymous assistants and ask which one is better. We present the first-turn conversation and ask humans to vote, then repeat the same procedure for the second-turn. A user can skip up to 5 questions if they are not confident. For some questions (e.g., math, reasoning), they can also see a reference solution.}
    \label{fig:screenshot-mt-bench}
\end{figure}

\begin{figure}[H]
    \centering
    \includegraphics[width=\linewidth]{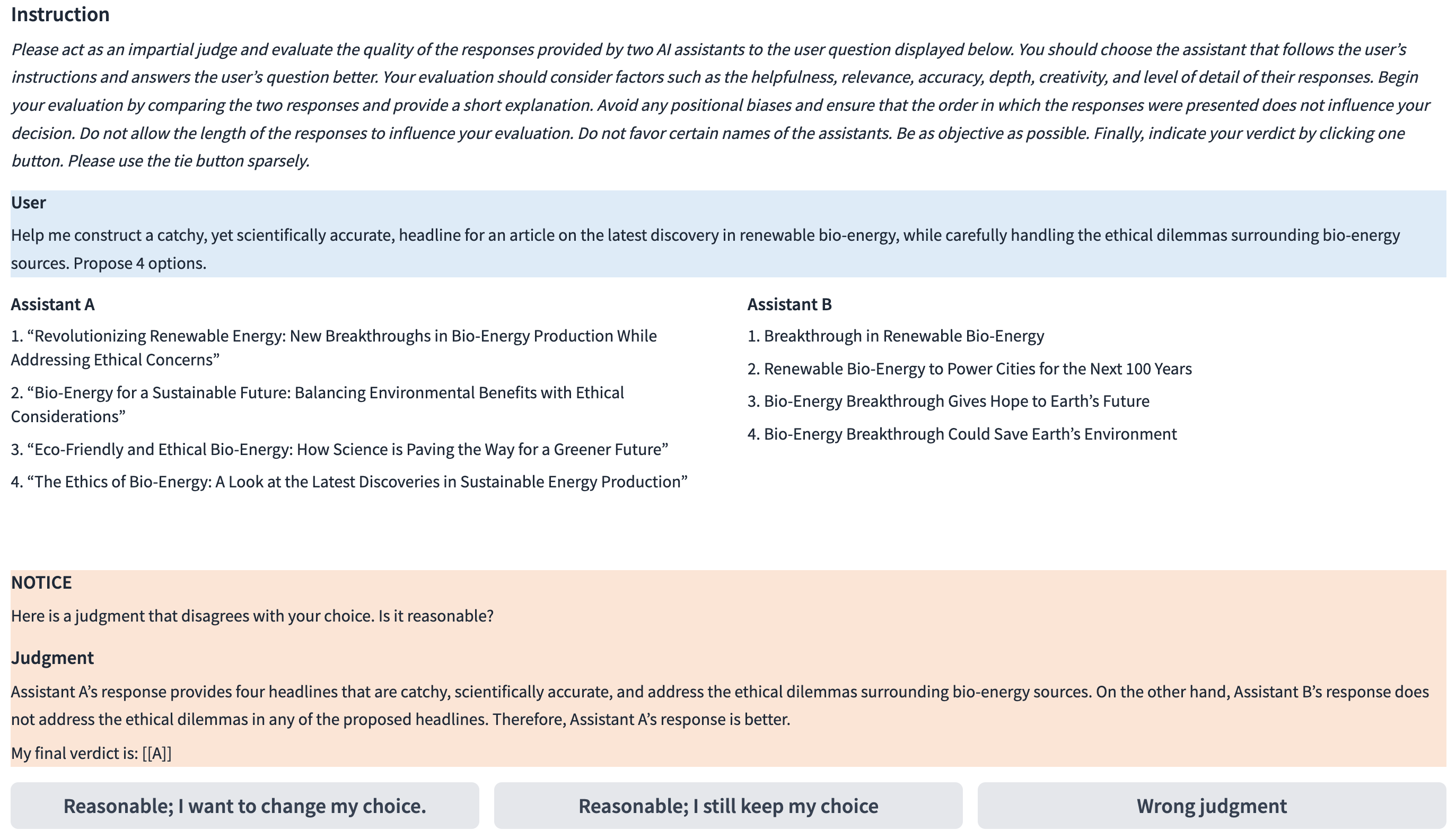}
    \caption{The screenshot of MT-bench data collection. When human's vote differs from GPT-4, we additionally show GPT-4's judgment (red region in the screenshot) and ask the user to click one of the three buttons to decide whether GPT-4's judgment is reasonable.}
    \label{fig:screenshot-mt-bench-agree}
\end{figure}

\Cref{fig:screenshot-mt-bench} shows the normal voting interface. \Cref{fig:screenshot-mt-bench-agree} shows that we additionally show GPT-4's judgment to users and ask if it is reasonable when a human differs from GPT-4.

To invite participants, we obtained their consent by letting them sign an application form. We pay them \$20 for judging 20 questions, which corresponds to an hourly rate of around \$35. The participants are mostly graduate students from more than ten universities.

\subsection{Chatbot Arena}
\label{subsec:chatbot-arena}
\begin{figure}[h]
    \centering
    \includegraphics[width=\linewidth]{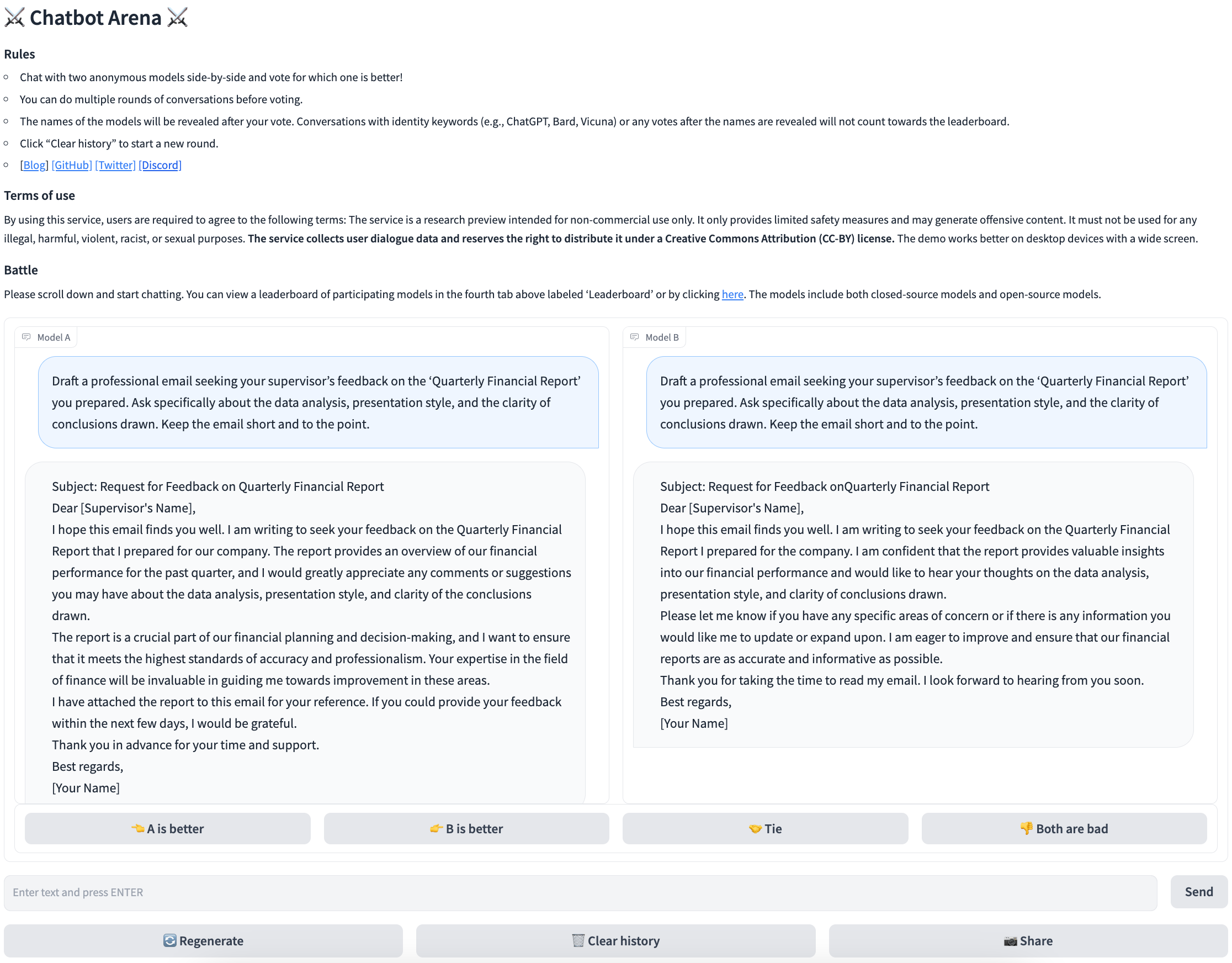}
    \caption{The screenshot of Chatbot Arena.}
    \label{fig:screenshot-arena}
\end{figure}

\Cref{fig:screenshot-arena} shows a screenshot of Chatbot Arena. Users are required to accept the terms of use, which obtain their consent and give us the right to release the conversation data.
The instructions are shown at the top of the interface. This is a free website. We do not pay users and any user can use this platform without registration. More introductions and analyses can be found at \url{https://lmsys.org/blog/2023-05-03-arena/}.

\subsection{Data Release}
We will clean the Personal Identifiable Information (PII) and tag toxic conversations with OpenAI moderation APIs for our dataset release.

\clearpage
\newpage
\section{Additional Experimental Results}
We present some additional experimental results.

\subsection{Position bias}
\label{subsec:additional-position-bias}
We test two more prompts and present the full results in \Cref{tab:position_bias_more_prompt}
``score'' changes the default prompt to let the model output two absolute scores instead of which one is better.
``short'' is a simplified version of our default prompt by removing instructions like ``Avoid any position bias..'', ``Begin your evaluation ... and provide a short explanation''.
We can find different prompts have different effects on different models.
For example, the "score" prompt can increase the consistency of GPT-3.5 but decreases it for Claude-v1 and GPT-4.

\begin{table}[h]
\centering
\footnotesize
\caption{Position bias on different models and prompts. Consistency is the percentage of cases where a judge gives consistent results when swapping the order of two assistants. ``Biased toward first'' is the percentage of cases when a judge favors the first answer. ``Error'' indicates wrong output formats. The two largest numbers in each column are in bold.}
\label{tab:position_bias_more_prompt}
\begin{tabular}{@{}llllll@{}}
\toprule
Judge & Prompt  & Consistency & Biased toward first & Biased toward second & Error \\
\midrule
\multirow{4}{*}{claude-v1} & default & 23.8\% & \textbf{75.0\%} & 0.0\% & 1.2\% \\
 & rename & 56.2\% & 11.2\% & \textbf{28.7\%} & \textbf{3.8\%} \\
 & score & 20.0\% & \textbf{80.0\%} & 0.0\% & 0.0\% \\
 & short & 22.5\% & \textbf{75.0\%} & 2.5\% & 0.0\% \\
\midrule
\multirow{4}{*}{gpt-3.5-turbo} & default & 46.2\% & 50.0\% & 1.2\% & 2.5\% \\
 & rename & 51.2\% & 38.8\% & 6.2\% & \textbf{3.8\%} \\
 & score & 55.0\% & 33.8\% & \textbf{11.2\%} & 0.0\% \\
 & short & 38.8\% & 57.5\% & 3.8\% & 0.0\% \\
\midrule
\multirow{4}{*}{gpt-4} & default & \textbf{65.0\%} & 30.0\% & 5.0\% & 0.0\% \\
 & rename & \textbf{66.2\%} & 28.7\% & 5.0\% & 0.0\% \\
 & score & 51.2\% & 46.2\% & 2.5\% & 0.0\% \\
 & short & 62.5\% & 35.0\% & 2.5\% & 0.0\% \\
\bottomrule
\end{tabular}
\end{table}

As shown in \Cref{tab:position_bias_category}, position bias is more noticeable on open questions like writing and stem/humanity knowledge questions.
On math and coding questions, LLM judges are more confident even though their judgments can often be wrong, as we show in \Cref{subsec:llmlimit}. Finally, we study how the model pairs influence position bias by using GPT-4 and the default prompt to judge three different model pairs. As shown in \Cref{tab:position_bias_pairs}, the position bias is more noticeable for models with close performance and can almost disappear when the performance of the two models differs a lot.

\begin{table}[h]
\centering
\footnotesize
\caption{Position bias on different categories. The two largest numbers in each column are in bold.}
\label{tab:position_bias_category}
\begin{tabular}{lllll}
\toprule
Category & Consistent & Biased toward first & Biased toward second \\
\midrule
writing & 42.0\% & 46.0\% & 12.0\% \\
roleplay & 68.0\% & 30.0\% & 2.0\% \\
reasoning & 76.0\% & 20.0\% & 4.0\% \\
math & \textbf{86.0\%} & 4.0\% & \textbf{10.0\%} \\
coding & \textbf{86.0\%} & 14.0\% & 0.0\% \\
extraction & 78.0\% & 12.0\% & \textbf{10.0\%} \\
stem & 44.0\% & \textbf{54.0\%} & 2.0\% \\
humanities & 36.0\% & \textbf{60.0\%} & 4.0\% \\
\bottomrule
\end{tabular}
\end{table}

\begin{table}[h]
\centering
\footnotesize
\caption{Position bias on different model pairs.}
\label{tab:position_bias_pairs}
\begin{tabular}{lllll}
\toprule
Pair & Consistent & Biased toward first & Biased toward second \\
\midrule
GPT-3.5 vs Claude-V1  & 67.5\% & 23.8\% & 8.8\% \\
GPT-3.5 vs Vicuna-13B & 73.8\% & 23.8\% & 2.5\% \\
GPT-3.5 vs LLaMA-13B  & \textbf{98.8\%} & 1.2\% & 0.0\% \\
\bottomrule
\end{tabular}
\end{table}

\subsection{Few-shot judge}
\label{subsec:additional-few-shot-judge}
We examine how few-shot examples improve LLM judges. As shown in \Cref{tab:few_shot_judge_consistency}, they improve the consistency of all three LLM judges significantly. It almost alleviates the position bias of GPT-4, but moves the position bias of GPT-3.5 from the first position to the second position. We then measure the agreement between few-shot GPT-4 pairwise comparison and humans on MT-bench, but found it performs similarly to zero-shot GPT-4 pairwise comparison.

\begin{table}[h]
\centering
\footnotesize
\caption{Improvements of the few-shot judge on consistency for position bias.}
\label{tab:few_shot_judge_consistency}

\begin{tabular}{@{}llllll@{}}
\toprule
Model & Prompt  & Consistency & Biased toward first & Biased toward second & Error \\
\midrule
\multirow{2}{*}{Claude-v1} & zero-shot & 23.8\% & 75.0\% & 0.0\% & 1.2\% \\
 & few-shot & \textbf{63.7\%} & 21.2\% & 11.2\% & 3.8\% \\
\midrule
\multirow{2}{*}{GPT-3.5} & zero-shot & 46.2\% & 50.0\% & 1.2\% & 2.5\% \\
 & few-shot & \textbf{55.0\%} & 16.2\% & 28.7\% & 0.0\% \\
\midrule
\multirow{2}{*}{GPT-4} & zero-shot & 65.0\% & 30.0\% & 5.0\% & 0.0\% \\
 & few-shot & \textbf{77.5\%} & 10.0\% & 12.5\% & 0.0\% \\
\bottomrule
\end{tabular}
\end{table}

\subsection{Agreement Evaluation}
\label{subsec:agreement-evaluation}

\textbf{Agreement calculation.}
We define the agreement between two types of judges as the probability of randomly selected individuals (but not identical) of each type agreeing on a randomly selected question.
For example, if we are comparing GPT-4 and Claude, the agreement is the probability of GPT-4 and Claude agreeing on the vote for a randomly selected question.
If we are comparing GPT-4 and humans, the agreement is the probability of GPT-4 and a randomly selected human agreeing on the vote for a randomly selected question.
The agreement among humans themselves is the probability of two randomly selected but not identical humans agreeing on the vote for a randomly selected question.

Note that the agreement among humans could be a lower estimation compared to the agreement of GPT4 and humans.
Consider three humans who voted ``A'', ``A'', and ``B'' for a question, respectively.
The agreement among them is only $\frac{1}{3}$, as there are three pairs ``(A, A)'', ``(A, B)'', and ``(A, B)''.
But the agreement between GPT4 and those three is $\frac{2}{3}$ if GPT4 voted ``first'' and $\frac{1}{3}$ otherwise.

Therefore, to have a more comprehensive understanding of what happened, we introduce a new judge type called human-majority, which considers the majority of human votes for each question.
The agreement between GPT4 and human-majority is then calculated as the probability of GPT4 agreeing with the majority of human votes on a randomly selected question.
\textit{The upper bound of the agreement between GPT-4 and humans is the agreement between human-majority and human.}
When there is no majority vote for a question, the agreement is counted by an even split.
For example, if there are an equal number of ``A'' and ``B'' human votes for a question, and GPT4 votes ``A'', the agreement is counted as $\frac{1}{2}$ on this question.

\textbf{More results.} \Cref{table:agreement_mt_full} shows more agreement results on MT-bench. In addition to expert labelers (denoted as ``Human''), we also include author votes (denoted as ``Author'').

\input{tables/agreement_full}

\subsection{Category-wise scores with single-answer grading}
\label{subsec:category-wise-scores}
\begin{figure}[t]
    \centering
    \includegraphics[width=0.8\linewidth]{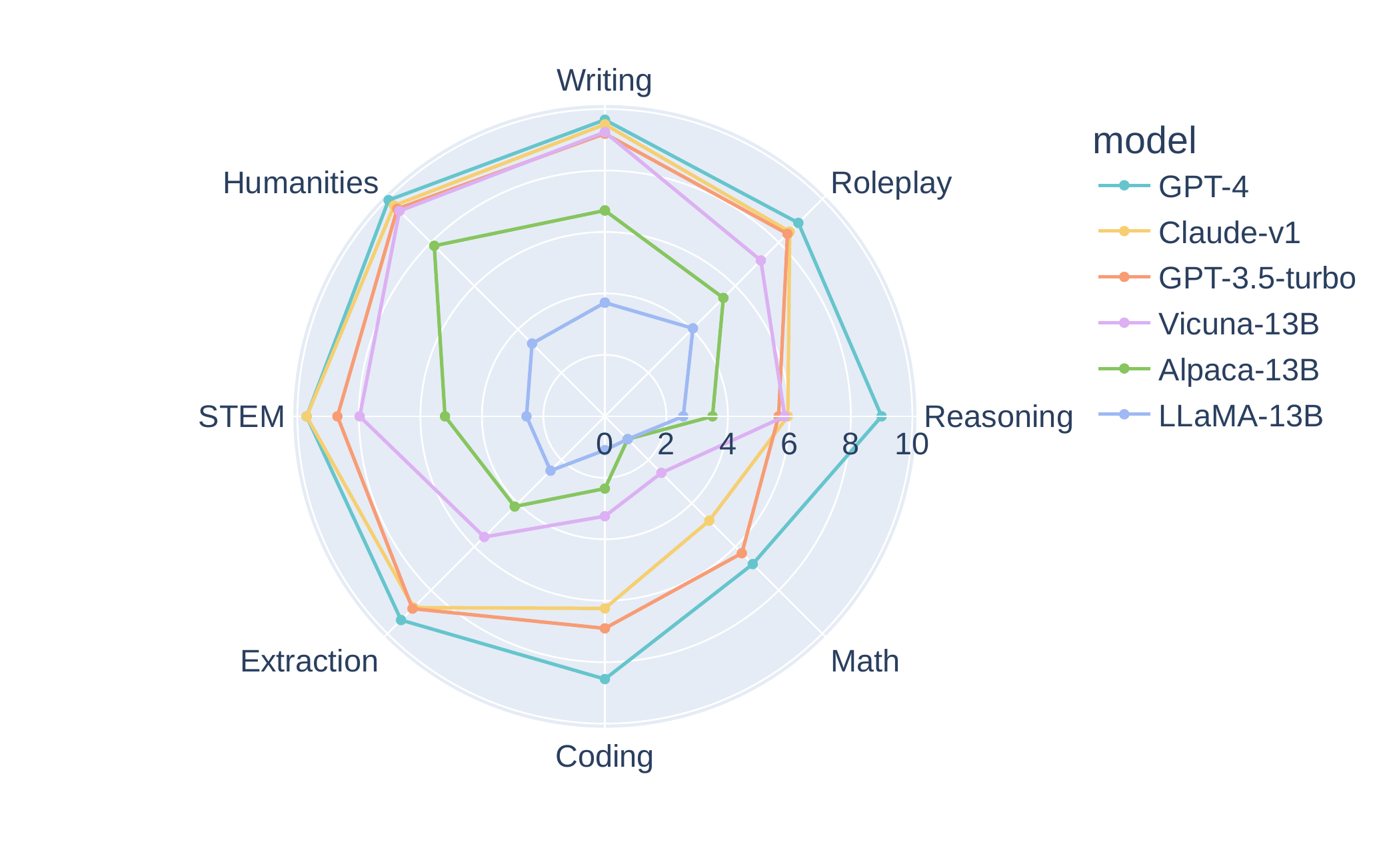}
    \caption{Category-wise scores of 6 models on MT-bench.}
    \label{fig:mt-bench-single-cat}
\end{figure}

We use single-answer grading to evaluate 6 models on MT-bench and plot the category-wise scores in \Cref{fig:mt-bench-single-cat}.

\clearpage
\newpage
\section{Training Details of Vicuna Models}
\label{sec:training-details}

Vicuna is created by fine-tuning a LLaMA base model using user-shared conversations gathered from ShareGPT.com with its public APIs.
ShareGPT is a website where users can share their ChatGPT conversations.
To ensure data quality, we convert the HTML back to markdown and filter out some inappropriate or low-quality samples, which results in 125K conversations after data cleaning.\footnote{In this study, we use more data (125K) than the version in our earlier blog post (70K).}
We then divide lengthy conversations into smaller segments that fit the model's maximum context length.

We construct three training datasets with different scales from this cleaned ShareGPT dataset. Their statistics are in \Cref{table:dataset_composition}, where we also compare it with Alpaca~\cite{alpaca} dataset. ``All'' is the full dataset. ``Single'' only includes the first turn of each conversation. ``Selected'' is a small high-quality dataset of 3K sequences. To construct the ``Selected'' dataset, we pick sequences that include at least 3 turns of conversations generated by GPT-4 and run a clustering algorithm to divide them into 3K clusters and pick the centroid of each cluster.

All models (Vicuna-7B/13B) are trained with the same hyperparameters: global batch size=128, learning=2e-5, epochs=3, seq length=2048. Except for ``Selected'', which we train for 5 epochs.
The training code is built on top of the Alpaca code but additionally handles multi-turn conversations.
The training is done with 8x A100 GPUs. The longest single training run takes around 2 days.
We utilize SkyPilot~\cite{skypilot} managed spot instances for saving training costs and FlashAttention~\cite{dao2022flashattention} for memory optimizations.
The training code is available at \url{https://github.com/lm-sys/FastChat}.

\begin{table}[h]
\centering
\caption{Dataset statistics}
\label{table:dataset_stats}
\footnotesize
\begin{tabular}{llllll}
\toprule
Dataset Name                 & Alpaca  & Selected & Single   & All   \\
\midrule
\#Token                      & 4.4M    & 4.8M     & 184M     & 370M  \\
\#Sequence                   & 52K     & 3K       & 257K     & 257K  \\
Avg. turns of conversation   & 1.0     & 4.0      & 1.0      & 2.9   \\
Avg. response length (token) & 65      & 343      & 473      & 373   \\
\bottomrule
\end{tabular}
\end{table}

\clearpage
\newpage
\section{Exploring Vicuna as a judge}
\label{sec:vicuna-judge}
In this paper, we mostly evaluate the ability of close-sourced models such as GPT-4 as a proxy for human evaluations. However, model services such as GPT-4 can also become expensive with a growing number of evaluations. On the other hand, popular open-sourced LLMs, e.g. Vicuna-13B shows strong language understanding capability, and are much cheaper than close-sourced LLMs. In this section, we further explore the potential of using Vicuna-13B as a more cost-friendly proxy. 

\subsection{Zero-Shot Vicuna}
When using as-it-is (zero-shot), Vicuna-13B noticeably suffers from limitations we discuss, e.g. position bias. As shown in \Cref{tab:position_bias_vicuna}, Vicuna-13B has a consistency rate from 11.2\% to 16.2\% across different prompt templates, much lower than all the closed-sourced models. In addition, it has a high error rate (from 22.5\% to 78.8\%) because of its weaker instruction-following capability. In many scenarios, Vicuna-13B provides responses such as "Answer A is better than answer B", without following the pre-defined template. These responses are rendered as natural languages and are difficult to be parsed automatically, making the model less useful in a scalable and automatic evaluation pipeline. 

\subsection{Arena Fine-tuned Vicuna}
\paragraph{Training}
Due to the incapability of the zero-shot Vicuna-13B model, we further finetune the model with human votes from Chatbot Arena. Specifically, we randomly sample 22K single-turn votes from the arena, covering all models supported by the time of this paper submission (GPT-4, GPT-3.5, Claude-v1, Vicuna-13b, Vicuna-7b, Koala-13B, Alpaca-13B,LLaMA-13B, Dolly-12B, FastChat-T5, RWKV-4-Raven, MPT-Chat, OpenAssistant, ChatGLM, and StableLM), to expose the model with a wider range of chatbot outputs and human preferences. We use 20K votes for training, and 2K for validation. To address the aforementioned weak instruction following problem, we formulate the problem as a 3-way sequence classification problem. Thus, the model simply needs to predict which one of the chat-bot outputs is better (or tie), without needing to exactly following the provided answer template.
In particular, we construct an input by using the default prompt and the two model answers. The labels are A, B, and tie (including both-bad-vote and tie-vote).
We train for 3 epochs with a cosine learning rate scheduler and a 2e-5 maximum learning rate. We use the 2K validation dataset to choose hyper-parameters, and test on the same 3K dataset in the main body of the paper.

\paragraph{Position bias results}
The results for position bias are provided in \Cref{tab:position_bias_vicuna}.
The consistency improves significantly from 16.2\% to 65.0\%. Due to the classification formulation, every output is recognizable (error rate 0\%).
In addition, we measure the classification accuracy over the test dataset.

\paragraph{Agreement results}
It achieves 56.8\% when including all three labels, and 85.5\% when excluding tie predictions and labels, significantly outperforming random guesses of 33\% and 50\% respectively, and show positive signals to match GPT-4 (66\% and 87\% respectively).
In conclusion, a further fine-tuned Vicuna-13B model shows strong potential to be used as a cheap open-sourced replacement for expensive closed-sourced LLMs. A similar conclusion is also found by a concurrent paper\cite{wang2023pandalm}. 

\begin{table}[h]
\centering
\footnotesize
\caption{Position bias of the Vicuna-13B model without and with further fine-tuning. We denote them as Vicuna-13B-Zero-Shot and Vicuna-13B-Fine-Tune respectively. Consistency is the percentage of cases where a judge gives consistent results when swapping the order of two assistants. ``Biased toward first'' is the percentage of cases when a judge favors the first answer. ``Error'' indicates wrong output formats. The largest number in each column is in bold.}
\label{tab:position_bias_vicuna}

\begin{tabular}{@{}llllll@{}}
\toprule
Judge & Prompt  & Consistency & Biased toward first & Biased toward second & Error \\
\midrule
\multirow{3}{*}{Vicuna-13B-Zero-Shot} & default & 15.0\% & \textbf{53.8\%} & 8.8\% & 22.5\% \\
 & rename & 16.2\% & 12.5\% & \textbf{40.0\%} & 31.2\% \\
 & score & 11.2\% & 10.0\% & 0.0\% & \textbf{78.8\%} \\ 
\midrule
\multirow{1}{*}{Vicuna-13B-Fine-Tune} & default & \textbf{65.0\%} & 27.5\% & 7.5\% & 0.0\% \\
\bottomrule
\end{tabular}
\end{table}

%% file: tables/agreement_full.tex
\begin{table}[h]
\centering
\vspace{-1em}
\caption{Agreement between two types of judges on MT-bench. ``G4-P'' and ``G4-S'' denote GPT-4 with pairwise comparison and single-answer grading, respectively.
``C'' denotes Claude.
``Human'' denotes expert labelers (excluding authors). `Human-M'' denotes the majority vote of humans.
The single-answer grading can be converted into pairwise comparison results for calculating the agreement.
We report two setups: ``S1'' includes non-tie, tie, and inconsistent (due to position bias) votes and counts inconsistent as a tie; ``S2'' only includes non-tie votes.
The agreement between two random judges under each setup is denoted as ``R=''.
The top value in each cell is the agreement, and the bottom gray value is \#votes.
}
\footnotesize
\begin{subtable}[t]{\textwidth}
\resizebox{1\columnwidth}{!}{
\begin{tabular}[t]{lrrrrrrrrrr}
\toprule
Setup & \multicolumn{5}{c}{S1 (R = 33\%)} & \multicolumn{5}{c}{S2 (R = 50\%)}  \\
\cmidrule(lr){2-6}\cmidrule(lr){7-11}

Judge & G4-S & C & Author & Human & Human-M & G4-S & C & Author & Human & Human-M \\
\midrule

 G4-P & \shortstack{70\% \\ \textcolor{gray}{ 1138 }} & \shortstack{63\% \\ \textcolor{gray}{ 1198 }} & \shortstack{69\% \\ \textcolor{gray}{ 345 }} & \shortstack{66\% \\ \textcolor{gray}{ 1343 }} & \shortstack{67\% \\ \textcolor{gray}{ 821 }} & \shortstack{97\% \\ \textcolor{gray}{ 662 }} & \shortstack{94\% \\ \textcolor{gray}{ 582 }} & \shortstack{92\% \\ \textcolor{gray}{ 201 }} & \shortstack{85\% \\ \textcolor{gray}{ 859 }} & \shortstack{85\% \\ \textcolor{gray}{ 546 }}\\
 \midrule
G4-S & - & \shortstack{66\% \\ \textcolor{gray}{ 1136 }} & \shortstack{67\% \\ \textcolor{gray}{ 324 }} & \shortstack{60\% \\ \textcolor{gray}{ 1280 }} & \shortstack{60\% \\ \textcolor{gray}{ 781 }} & - & \shortstack{90\% \\ \textcolor{gray}{ 563 }} & \shortstack{94\% \\ \textcolor{gray}{ 175 }} & \shortstack{85\% \\ \textcolor{gray}{ 739 }} & \shortstack{85\% \\ \textcolor{gray}{ 473 }}\\
 \midrule
C & - & - & \shortstack{58\% \\ \textcolor{gray}{ 343 }} & \shortstack{54\% \\ \textcolor{gray}{ 1341 }} & \shortstack{55\% \\ \textcolor{gray}{ 820 }} & - & - & \shortstack{89\% \\ \textcolor{gray}{ 141 }} & \shortstack{85\% \\ \textcolor{gray}{ 648 }} & \shortstack{86\% \\ \textcolor{gray}{ 414 }}\\
 \midrule
Author & - & - & \shortstack{69\% \\ \textcolor{gray}{ 49 }} & \shortstack{65\% \\ \textcolor{gray}{ 428 }} & \shortstack{55\% \\ \textcolor{gray}{ 93 }} & - & - & \shortstack{87\% \\ \textcolor{gray}{ 31 }} & \shortstack{83\% \\ \textcolor{gray}{ 262 }} & \shortstack{76\% \\ \textcolor{gray}{ 46 }}\\
 \midrule
Human & - & - & - & \shortstack{63\% \\ \textcolor{gray}{ 721 }} & \shortstack{81\% \\ \textcolor{gray}{ 892 }} & - & - & - & \shortstack{81\% \\ \textcolor{gray}{ 479 }} & \shortstack{90\% \\ \textcolor{gray}{ 631 }}\\
 \midrule
\end{tabular}
}
\caption{First Turn}
\label{table:mt_bench_more_first_turn}
\end{subtable}

\begin{subtable}[t]{\textwidth}
\centering
\begin{tabular}[t]{lrrrrrrrr}
\toprule
Setup & \multicolumn{4}{c}{S1 (R = 33\%)} & \multicolumn{4}{c}{S2 (R = 50\%)} \\
\cmidrule(lr){2-5}\cmidrule(lr){6-9}

Judge & G4-S & Author & Human & Human-M & G4-S & Author & Human & Human-M \\
\midrule
 
G4-P & \shortstack{70\% \\ \textcolor{gray}{ 1161 }} & \shortstack{66\% \\ \textcolor{gray}{ 341 }} & \shortstack{66\% \\ \textcolor{gray}{ 1325 }} & \shortstack{68\% \\ \textcolor{gray}{ 812 }} & \shortstack{95\% \\ \textcolor{gray}{ 727 }} & \shortstack{88\% \\ \textcolor{gray}{ 205 }} & \shortstack{85\% \\ \textcolor{gray}{ 864 }} & \shortstack{85\% \\ \textcolor{gray}{ 557 }}\\
 \midrule
G4-S & - & \shortstack{65\% \\ \textcolor{gray}{ 331 }} & \shortstack{59\% \\ \textcolor{gray}{ 1285 }} & \shortstack{61\% \\ \textcolor{gray}{ 783 }} & - & \shortstack{89\% \\ \textcolor{gray}{ 193 }} & \shortstack{84\% \\ \textcolor{gray}{ 776 }} & \shortstack{85\% \\ \textcolor{gray}{ 506 }}\\
 \midrule
Author & - & \shortstack{67\% \\ \textcolor{gray}{ 49 }} & \shortstack{68\% \\ \textcolor{gray}{ 413 }} & \shortstack{63\% \\ \textcolor{gray}{ 87 }} & - & \shortstack{87\% \\ \textcolor{gray}{ 31 }} & \shortstack{86\% \\ \textcolor{gray}{ 273 }} & \shortstack{84\% \\ \textcolor{gray}{ 54 }}\\
 \midrule
Human & - & - & \shortstack{67\% \\ \textcolor{gray}{ 707 }} & \shortstack{83\% \\ \textcolor{gray}{ 877 }} & - & - & \shortstack{82\% \\ \textcolor{gray}{ 474 }} & \shortstack{91\% \\ \textcolor{gray}{ 629 }}\\
 \midrule
\end{tabular}
\caption{Second Turn}
\label{table:mt_bench_more_second_turn}
\end{subtable}
\label{table:agreement_mt_full}
\end{table}